\documentclass[10pt,twocolumn,letterpaper]{article}

\usepackage{wacv}
\usepackage{times}
\usepackage{epsfig}
\usepackage{graphicx}
\usepackage{amsmath}
\usepackage{amssymb}
\usepackage{pdfpages}
\usepackage{multido}
\usepackage[pagebackref=true,breaklinks=true,letterpaper=true,colorlinks,bookmarks=false]{hyperref}
\usepackage{comment}
\usepackage[dvipsnames]{xcolor}
\usepackage{enumitem}
\usepackage{array}
\usepackage{varwidth}
\usepackage[ruled,vlined]{algorithm2e}
\usepackage{cases}

\usepackage[accsupp]{axessibility}

\newcommand{\domainS}{$\mathcal{D_S}$}
\newcommand{\domainT}{$\mathcal{D_T}$}

\newcommand{\algoname}{D4}
\newcommand{\gtacs}{GTA5$\rightarrow$CS}
\newcommand{\synthiacs}{SYNSEQ$\rightarrow$CS}

\SetKwComment{Comment}{}{}

\def\wacvPaperID{430} % Enter the WACV Paper ID here

\wacvfinalcopy % *** Uncomment this line for the final submission

\ifwacvfinal
\usepackage[breaklinks=true,bookmarks=false]{hyperref}
\else
\usepackage[pagebackref=true,breaklinks=true,colorlinks,bookmarks=false]{hyperref}
\fi

\ifwacvfinal
\else
\fi

\begin{document}

%%%%%%%%% TITLE
\title{Plugging Self-Supervised Monocular Depth into \\
Unsupervised Domain Adaptation for Semantic Segmentation}

\author{Adriano Cardace\quad Luca De Luigi\quad Pierluigi Zama Ramirez\quad Samuele Salti\quad Luigi Di Stefano\\
Department of Computer Science and Engineering (DISI)\\
University of Bologna, Italy\\
{\tt\small \{adriano.cardace2, luca.deluigi4, pierluigi.zama\}@unibo.it}
% \and
}

\maketitle
\thispagestyle{empty}

%%%%%%%%% ABSTRACT
\begin{abstract}
    Although recent semantic segmentation methods have made remarkable progress, they still rely on large amounts of annotated training data, which are often infeasible to collect in the autonomous driving scenario. Previous works usually tackle this issue with Unsupervised Domain Adaptation (UDA), which entails training a network on synthetic images and applying the model to real ones while minimizing the discrepancy between the two domains. Yet, these techniques do not consider additional information that may be obtained from other tasks. Differently, we propose to exploit self-supervised monocular depth estimation to improve UDA for semantic segmentation. On one hand, we deploy depth to realize a plug-in component which can inject  complementary  geometric cues into any existing UDA method. We further rely on depth to generate a large and varied set of samples to Self-Train the final model.
    Our whole proposal allows for achieving state-of-the-art performance (58.8 mIoU) in the \gtacs{} benchmark.
    Code is available at {\small{\url{https://github.com/CVLAB-Unibo/d4-dbst}}}.
\end{abstract}

\section{Introduction}
\label{sec:intro}
Semantic segmentation is the task of classifying each pixel of an image. Nowadays, Convolutional Neural Networks can achieve impressive results in this task but require huge quantities of labelled images at training time \cite{cnnss, deeplabv2, Enet, unet}. A popular trend to address this issue concerns leveraging on computer graphics simulations \cite{synthia} or game engines \cite{gta} to obtain automatically synthetic images endowed with per-pixel semantic labels. Yet, a network trained on synthetic data only will perform poorly in real environments due to the so called \emph{domain-shift} problem. In the last few years, many Unsupervised Domain Adaptation (UDA) techniques aimed at alleviating the domain-shift problem have been proposed in literature. These approaches try to minimize the gap between the labeled source domain (e.g. synthetic images) and the unlabeled target domain (e.g. real images) by either hallucinating input images, manipulating the learned features space or imposing statistical constraints on the predictions \cite{dcan, road, Zhang_2017, hoffman2016fcns}.

\begin{figure}[t]
    \centering
    \includegraphics[width=\linewidth]{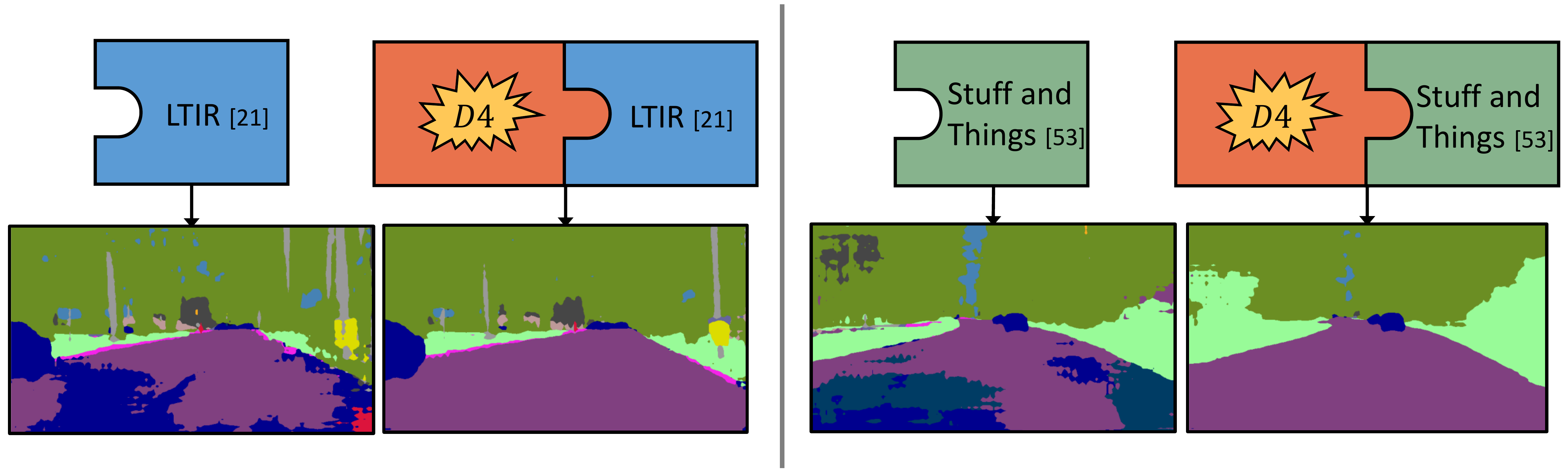}
    \caption{\algoname{} can be plugged seamlessly into any existing method to improve UDA for Semantic Segmentation.  Here we show how the introduction of \algoname{} can ameliorate the performance of two recent methods like  LTIR \cite{ltir} and Stuff and Things \cite{stuffAndThings}.}
    \label{fig:teaser}
\end{figure}

At a more abstract level, UDA may be thought of as the process of transferring more effectively to the target domain the knowledge from a task solved in the source domain. This suggests that it may be possible to improve UDA by transferring also knowledge learned from \emph{another task} to improve performance in the real domain. In fact, the existence of tightly related representations within CNNs trained for different tasks has been highlighted since the early works in the field \cite{transferable}, and it is nowadays standard practice to initialize CNNs deployed for a variety of diverse tasks, such as, e.g., object detection \cite{efficientdet}, semantic segmentation \cite{deeplabv3+} and monocular depth estimation \cite{monodepth2}, with weights learned on Imagenet  Classification  \cite{imagenet}. The notion of \textit{transferability} of representations among CNNs trained to solve different visual tasks has been formalized computationally by the Taskonomy proposed  in \cite{taskonomy}. Later, \cite{atdt} has shown that it is possible to train a CNN to hallucinate deep features learned to address one task into features amenable to another task related to the former.

Inspired by these findings, we argue that  monocular depth estimation could be an excellent task in order to gather additional knowledge useful to address semantic segmentation in UDA settings.
First of all, a monocular depth estimation network makes predictions based on 
3D cues dealing with the appearance, shape,  relative sizes and spatial relationships of the stuff and things observed in the training images. This suggests that the network has to predict geometry by implicitly learning to understand  the semantics of the scene. Indeed, \cite{geometry, semantic_booster,  Semantic_Guidance, semanticallyguided} show that a monocular depth estimation network obtains better performances if forced to learn jointly a semantic segmentation task. We argue, though, the correlation between geometry and semantics to hold bidirectionally, such that a semantic segmentation network may obtain useful hints from depth information. This intuition is supported by \cite{atdt}, which shows that it is possible to learn a mapping in both directions between features learned to predict depth and per-pixel semantic labels.
It is also worth observing how depth prediction networks tend to extract accurate information for  regions characterized by repeatable and simple  geometries, such as roads and buildings, which feature strong spatial and geometric priors (e.g. the road is typically a plane in the bottom part of the image) \cite{monodepth, monodepth2, monoresmatch, depthhints}. Therefore, on one hand predicting accurately the semantics of such regions from depth information alone should be possible. On the other, a semantic network capable of reasoning on the geometry of the scene should be less prone to mistakes caused by appearance variations between synthetic and real images, the key issue in UDA for semantic segmentation.

Despite the above observations, injection of geometric cues into UDA frameworks for semantic segmentation has been largely unexplored in literature, with the exception of a few proposals, which either assume availability of depth labels in the real domain \cite{multichannel}, a very restrictive assumption, or can leverage on depth information only in the synthetic domain due to availability of cheap labels \cite{dada, spigan, geoguided}. In this respect, we set forth  an additional consideration: nowadays effective self-supervised procedures allow for training a monocular depth estimation network without the need of ground-truth labels \cite{monodepth2, Garg_2016, Zhou_2017}.

Based on the above intuitions and considerations, in this paper we propose the first approach that, thanks to self-supervision, allows for deploying depth information from both synthetic and \emph{unlabelled real} images in order to inject geometric cues in UDA for semantic segmentation. Purposely, we  adapt the knowledge learned to pursue depth estimation into a representation amenable to semantic segmentation by the feature transfer architecture proposed in \cite{atdt}. As the geometric cues learned from monocular images yield semantic predictions that are often complementary to those attainable by current UDA methods, we realize our proposal as a depth-based add-on, dubbed \algoname{} (Depth For), which can be plugged seamlessly into any UDA method to boost its performances, as illustrated in Fig. \ref{fig:teaser}.  

A recent trend in UDA for semantic segmentation is Self-Training (ST), which consists  in further fine-tuning the trained network by  its own predictions \cite{cbst, crst, rectifying, pycda, Pan_2020, instance_adaptive}. We propose a novel Depth-Based Self-Training (DBST) approach which deploys once more  the availability of depth information  for real images in order to build a large and varied dataset of plausible samples to be deployed in the Self-Training (ST) procedure\footnote{See also \cite{hoyer2021three} for concurrent work that proposes a similar idea.}.

Our framework can improve many state-of-the-art methods by a large margin in two UDA for semantic segmentation  benchmarks, where networks are trained either on GTA5~\cite{gta} or SYNTHIA VIDEO SEQUENCES~\cite{synthia} and tested on Cityscapes~\cite{Cityscapes}. Moreover, we show that our DBST procedure enables to distill the whole framework into a single ResNet101~\cite{resnet} and achieve state-of-the-art performance. Our contributions can be summarized as follows:
\begin{itemize}
    \item We are the first to show how to exploit self-supervised monocular depth estimation on real images to pursue semantic segmentation in a domain adaptation settings. 
    
    \item We propose a depth-based module (\algoname{}) which can be plugged into any UDA for semantic segmentation method to boost performance.  

    \item We introduce a new protocol (DBST) that exploits depth predictions to synthesize augmented training samples for the final self-training step deployed oftentimes in UDA for semantic segmentation pipelines.
    
    \item  We show that leveraging on both \algoname{} and DBST allows for achieving 58.8 mIoU in the popular \gtacs{} UDA benchmark, i.e., to the best of our knowledge, the new state-of-the-art. 
\end{itemize}

\section{Related Work}

\textbf{Domain Adaptation.}
Domain Adaptation is a promising way of solving semantic segmentation without annotations.
Pioneering works \cite{cycada, dcan, Chang_2019, Choi_2019, Murez_2018, Zhang_2018, ipas, bdl, ltir} rely on CycleGANs \cite{cyclegan} to convert source data into the style of target data, reducing the low-level visual appearance discrepancy among domains.
Other works exploit adversarial training to enforce domain alignment \cite{adaptsegnet, patch, fada, joint_adversarial, adversarial_perturbation, pizzati, michieli, ADVENT}. \cite{stuffAndThings} extended this idea by aligning differently objects with low and high variability in terms of appearance.
Few works tried to exploit depth information to boost UDA for semantic segmentation. \cite{dada}, for example, proposes a unified depth-aware UDA framework that leverages the knowledge of depth maps in the source domain to perform feature space alignment. \cite{saha2021learning} extends this idea by modelling  explicitly the relation between different visual semantic classes and depth ranges. \cite{inout_adaptation}, instead, considers depth as a way to obtain adaptation at both the input and output level. \cite{multichannel} is the first work to consider depth in the target domain, although assuming supervision to be available. Conversely, we show how to deploy depth in the target domain without availability of ground-truth depths.

\textbf{Self-Training.} More recently, a new line of research focuses on self-training~\cite{pseudolabel}, where a semantic classifier is fine-tuned directly on the target domain, using its own predictions as pseudo-labels. \cite{cbst, crst, instance_adaptive} cleverly set class-confidence thresholds to mask wrong predictions. 
\cite{mrnet, Pan_2020, rectifying} propose to use pseudo-labels with different regularization techniques to minimize both the inter-domain and intra-domain gap.
\cite{zhang2021prototypical} instead, estimates the likelihood of pseudo-labels to perform online correction and denoising during training.
Differentely, \cite{dacs} synthesizes new samples for the target domain by cropping objects from source images using ground truth labels and pasting them onto target images. Inspired by this work, we propose a novel algorithm for generating new samples to perform self-training on the target domain. In contrast to \cite{dacs}, our strategy is applied to target images only and relies on the availability of depth maps obtained through self-supervision. 

\textbf{Task Adaptation.} 
All existing approaches tackle independently task adaptation or domain adaptation. \cite{tzeng2015simultaneous} was the first paper to propose a cross-tasks and cross-domains adaptation approach, considering two image classification problems as different tasks.
UM-Adapt~\cite{kundu2019adapt} employs a cross-task distillation module to force inter-task coherency. % while minimizing the domain discrepancy with a custom cross-task loss.
Differently, \cite{atdt}, directly exploits the relationship among tasks to reduce the need for labelled data. This is done by learning a mapping function in feature space between two networks trained independently for two separate tasks, a pretext and target one. We leverage on this intuition but, unlike \cite{atdt}, our approach does not require  supervision to solve the pretext task in the target domain.

\begin{figure}[t]
    \centering
    \includegraphics[width=\linewidth]{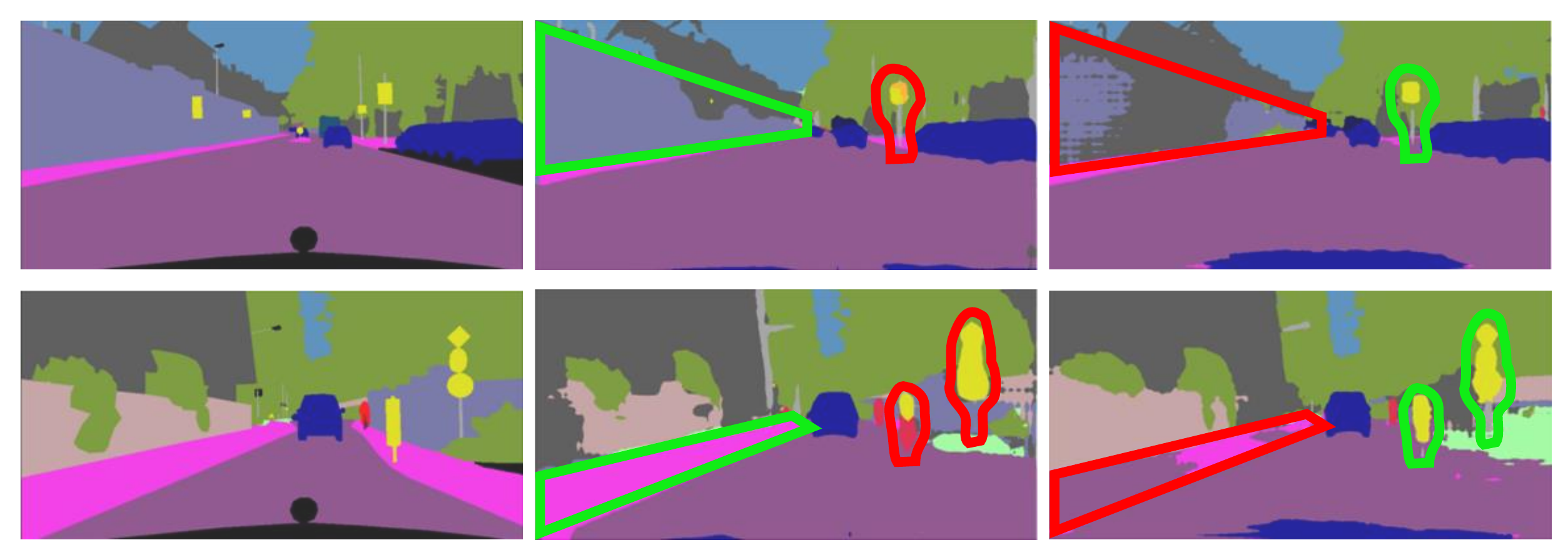}
    \caption{From left to right: ground truth, semantics from depth, semantics by LTIR~\cite{ltir}. The semantic labels predicted from depth are more accurate than those yielded by UDA methods in regularly-shaped objects (such as the \textit{wall} in the top image and the \textit{sidewalk} in the bottom one), whilst UDA approaches tend to perform better on small objects (see the \textit{traffic signs} in both images).}
    \label{fig:davsta}
\end{figure}

\begin{figure*}[t]
    \centering
    \includegraphics[width=0.7\linewidth]{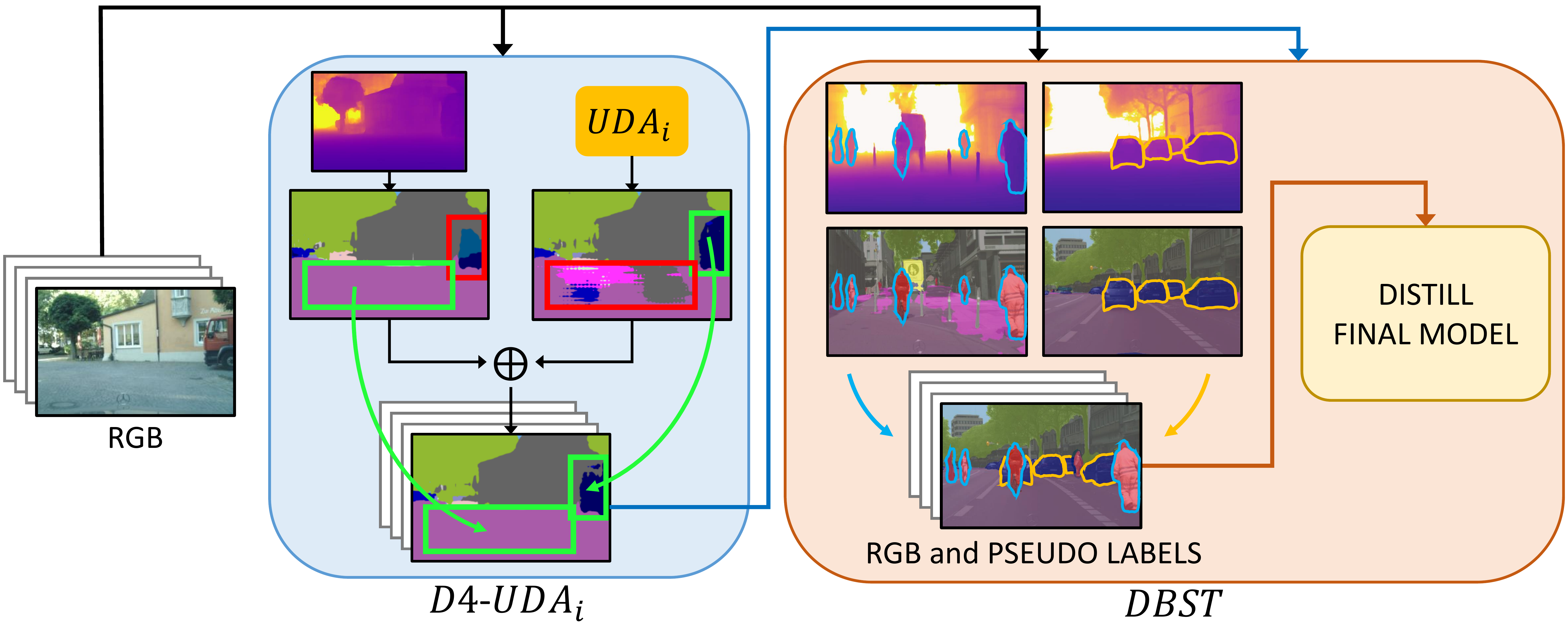}
    \caption{Overview of our proposal. RGB images are first processed by two different segmentation engines to produce complementary predictions that are then combined by a weighted sum  which accounts for the relative strengths of the two engines (Eq. \ref{eq:merge}). During the next step (DBST), predictions from \algoname{}-UDA$_i$ are used to synthesize augmented samples by mixing portions of different images according to depth and semantics. The augmented samples are used to train a final model, so as to distill the whole pipeline into a single network.}
    \label{fig:pipeline}
\end{figure*}

\section{Method}
In Unsupervised Domain Adaptation (UDA) for semantic segmentation one wishes to solve semantic segmentation in a target domain, \domainT{}, though labels are available only in  another domain, referred to as source domain \domainS{}. In the following we describe the two ingredients of our proposal to better tackle this problem. In Sec. \ref{sec:d2s} we show how to transfer information from self-supervised monocular depth to semantic segmentation and merge this knowledge with any UDA method (\algoname{}-UDA, Depth For UDA). Then, in Sec. \ref{sec:selftrain} we introduce a  \underline{D}epth-\underline{B}ased \underline{S}elf-\underline{T}raining strategy (DBST) to further improve semantic predictions while distilling the whole framework into a single CNN.

\subsection{\algoname{} (Depth For UDA)}
\label{sec:d2s}
\textbf{Semantics from Depth.}
% The main intuition behind our work is that depth information can improve semantic segmentation in a UDA settings. Thus, we leverage on the synergy between two tasks in a cross-domain scenario.
The main intuition behind our work is that 
%depth information can improve semantic segmentation in a UDA setting. More specifically, we conjecture that 
semantic segmentation masks obtained exploiting depth information have peculiar properties that make them suitable to improve segmentation masks obtained with standard UDA methods. However, predicting semantics from depth is an arduous task. Indeed, we experiment several alternatives (see Sec. \ref{sec:Analysis} \textit{Alternative strategies to exploit depth}) and find out that the most effective way is a procedure similar to the one proposed in \cite{atdt}, which we adapt to the UDA scenario.
% This  problem has been effectively addressed in   by the idea of \textit{transferring} deep features between CNNs specialized toward the given tasks.
The pipeline works as follows: train one CNN to solve a first task on  \domainS{} and \domainT{}, train another CNN to solve a second task  on \domainS{} only (i.e. the only domain where ground truth labels for the second task are available) and, finally, train a \textit{transfer} function to map deep features extracted by the first CNN into deep features amenable to the second one. As the second CNN has been trained only on \domainS{}, also the transfer function can be trained only on \domainS{} but, interestingly, it can generalize to \domainT{}. As a consequence, at inference time one can solve the second task in \domainT{} based on the features transferred from the first task. We refer to \cite{atdt} for further details.

Hence, if we assume the first and second task to consist in depth estimation and semantic segmentation, respectively, the idea of transferring features
might be deployed in a UDA scenario since it gives the possibility to solve the second task on \domainT{} without the need of ground truth labels. However, the learning framework delineated in \cite{atdt} assumes  availability of ground-truth labels for the first task (depth estimation in our setting) also in \domainT{} (real images). As pointed out in Sec. \ref{sec:intro}, this assumption does not comply with the standard UDA for semantic segmentation problem formulation, which requires availability of semantic labels for source images (\domainS{}) alongside with unlabelled target images only (\domainT{}). To address this issue we propose to rely on \textit{depth proxy-labels} attainable from images belonging to both  \domainS{} and \domainT{} without the need of any ground-truth information. In particular, we propose to deploy one of the recently proposed deep neural networks, such as \cite{monodepth2}, that can be trained to perform monocular depth estimation based on a self-supervised loss that requires availability of raw image sequences only, i.e. without ground-truth depth labels. Thus, in our method we introduce the following protocol. First, we train a self-supervised monocular depth estimation network on both \domainS{} and \domainT{}. Then, we use this network to generate \textit{depth proxy-labels} for both domains. We point out that we use such network as an off-the-shelf algorithm without the aim of improving depth estimation. Finally, according to \cite{atdt}, we  train a first CNN to predict depth from images on both domains by the previously computed \textit{depth proxy-labels}, a second CNN to predict semantic labels on \domainS{} and a transfer network which allows for predicting semantic labels from depth features in \domainT{}. In the following, we will refer to such predictions as ``semantics from depth'' because they concern semantic information extracted from features amenable to perform monocular depth estimation.

\textbf{Combine with UDA.}
Fig. \ref{fig:davsta} compares semantic predictions obtained from depth by the protocol described in the previous sub-section and from a recent UDA method. The reader may observe a clear pattern: predictions from depth tend to be smoother and more accurate on objects with large and regular shapes, like \textit{road}, \textit{sidewalk}, \textit{wall} and \textit{building}. However, they turn out often imprecise in regions where depth predictions are less informative, like thin things partially overlapped with other objects or fine-grained structures in the background. As UDA methods tend to perform better on such classes (see Fig. \ref{fig:davsta}), our \algoname{} approach is designed to \textit{combine} the semantic knowledge extracted from depth with that provided by any chosen UDA method in order to achieve more accurate semantic predictions.
Depth information helps on large objects with regular shapes, which usually account for the majority of pixels in an image. On the contrary, UDA methods perform well in predicting semantic labels for categories that typically concern  much smaller fractions of the total number of pixel in an image, like e.g. the \textit{traffic signs} in Fig. \ref{fig:davsta}. 
This orthogonality suggests that a simple yet effective way to combine the semantic knowledge drawn from depth with that provided by UDA methods consists in a weighted sum of predictions, with weights computed according to the frequency of  classes in \domainS{}  (the domain where semantic labels are available). As weights given to UDA predictions ($\textbf{w}_{uda}$) should be larger for rarer classes, they can be computed as:
\begin{equation}
\label{eq:wda}
    \resizebox{0.37\hsize}{!}{$\textbf{w}_{uda} = [w_{uda}^1, \dots, w_{uda}^C]$} \quad
    \text{where} \quad
    w_{uda}^i = \frac{1}{ln(\delta + f^i)}
\end{equation}
where $C$ denotes the number of classes and $f^i = \frac{n^i}{n^{tot}}$ denotes their frequencies at the pixel level, \ie the ratio between the number $n^i$ of pixels labelled with class $i$ in \domainS{} and the total number $n^{tot}$ of labelled pixels in \domainS{}. 
Eq. \ref{eq:wda} is the standard formulation introduced in \cite{Enet} to  compute bounded weights inversely proportional to the frequency of classes. We set $\delta$ in Eq. \ref{eq:wda} to 1.02, akin to \cite{Enet}.% and many following works.

Accordingly, weights applied to semantic predictions drawn from depth ($\textbf{w}_{dep}$) are given by: 
\begin{equation}
\label{eq:wta}
    \textbf{w}_{dep} = [w_{dep}^1, ..., w_{dep}^C] \quad \text{where} \quad w_{dep}^i = 1 - w_{uda}^i.
\end{equation}

Thus, at each pixel of a given image we propose to combine semantics from depth and predictions yielded by any chosen UDA method as follows:
\begin{equation}
\label{eq:merge}
    \widehat{\textbf{y}}_f = \textbf{w}_{dep} \cdot \phi_T(\Tilde{\textbf{y}}_{dep}) + \textbf{w}_{uda} \cdot \phi_T(\Tilde{\textbf{y}}_{uda}),
\end{equation}
where $\widehat{\textbf{y}}_f$ is the final prediction, $\Tilde{\textbf{y}}_{dep}$ and $\Tilde{\textbf{y}}_{uda}$ are the logits associated with semantics from depth and the selected UDA method, respectively, $\phi_T$ denotes the \textit{softmax} function with a temperature term $T$ that we set to 6 in our experiments.

As illustrated in Fig. \ref{fig:pipeline}, the formulation presented in Eq. \ref{eq:merge} and symbolized as $\bigoplus$ can be used seamlessly to plug semantics obtained from self-supervised monocular depth into any existing UDA method. We will refer to the combination of a given UDA method with our \algoname{} with the expression \algoname{}-UDA. Experimental results (Sec. \ref{sec:results}) show that all recent s.o.t.a. UDA methods do benefit significantly from the complementary geometric cues brought in  by \algoname.

\begin{figure}[t]
    \centering
    \includegraphics[width=0.5\linewidth]{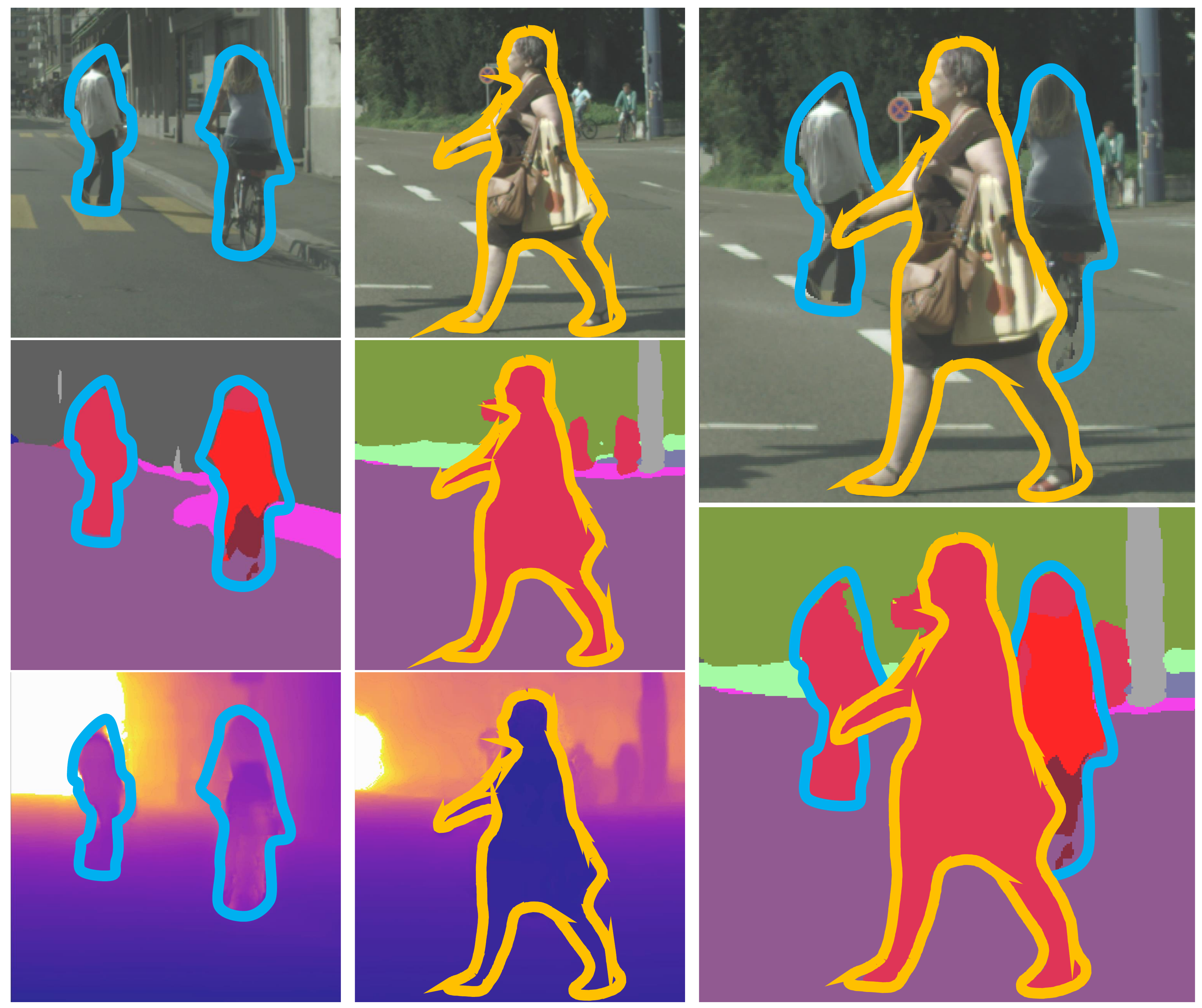}
    \caption{The rightmost column is synthesized by copying pixels from the left column into the central one. Pixels are chosen according to their semantic class (second row) and stacked according to their depths (third row). The white pixels in the depth maps represent areas too far from the camera that cannot be selected.}
    \label{fig:dbst}
\end{figure}

\subsection{DBST (Depth-Based Self-Training)}
\label{sec:selftrain}
We describe here our proposal to further improve semantic predictions and distill the knowledge of the entire system into a single network easily deployable at inference time. First, we predict semantic labels for every image in \domainT{} by our whole framework (i.e. \algoname{} alongside a selected UDA method, referred to as \algoname{}-UDA); then, we use these labels to train a new model on \domainT{}. This procedure, also known as self-training~\cite{pseudolabel}, has become popular in recent UDA for semantic segmentation literature \cite{cbst, crst, rectifying, pycda, Pan_2020, instance_adaptive} and consists in training a model by its own predictions, referred to as \textit{pseudo-labels}, sometimes through multiple iterations. On the other hand, we only perform one iteration, and the novelty of our approach concerns the peculiar ability to leverage on the depth information available for the images in \domainT{} to generate plausible new samples.

Running \algoname{}-UDA on \domainT{} yields semantic pseudo-labels for every image in \domainT{}. Yet, as described in Sec. \ref{sec:d2s} (\textit{Semantics from Depth}), each image in \domainT{} is also endowed with a depth prediction, provided by a self-supervised monocular depth estimation network.
We can take advantage of this information to formulate a novel depth-aware data augmentation strategy whereby portion of images and corresponding pseudo-labels are \textit{copied} onto others so as to synthesize samples for the self-training procedure. The crucial difference between similar approaches presented in  \cite{classmix,dacs} and ours consists in the deployment of depth information to steer the data augmentation procedure towards more plausible samples. Indeed, a first intuition behind our method deals with semantic predictions being less accurate for objects distant from the camera: as such predictions play the role of labels in self-training, we prefer to pick closer rather than distant objects in order to generate training samples. Moreover, we reckon certain kinds of objects, like persons, vehicles and traffic signs, to be more plausibly transferable across different images as they tend to be small and less bound to specific spatial locations. On the contrary, it is quite unlikely to merge seamlessly a piece of road or building from a given image into a different one.

Given $N$ randomly selected images $x^n$ from \domainT{}, with $n \in \{1,\dots,N\}$, paired with semantic pseudo-labels $s^n$ and depth predictions $d^n$, we augment $x^{1}$, by copying on it pixels from the set $\mathcal{X}^{src} = \{x^2,\cdots,x^N\}$. For each pixel of the augmented image we have $N$ possible candidates, one from $x^{1}$ itself and $N - 1$ from the images in $\mathcal{X}^{src}$. We filter such candidates according to two criteria: the predicted depth should be lower than a threshold $t$ and the semantic prediction should belong to a predefined set of classes, $C^*$. Hence, we define the set of depths of the filtered candidates at each spatial location $p$ as:
\begin{equation}
\label{eq:depth_candidates}
    D_{p} = \{ d_{p}^n\;|\;d_{p}^n < t \land s_{p}^n \in C^{*}\} \quad n \in \{1,\dots,N\}.
\end{equation}
In our experiments, for each image the depth threshold $t$ is set to the $80^{th}$ percentile of the depth distribution, so as to avoid selecting pixels from the farthest objects in the scene. $C^*$ contains all \textit{things} classes (e.g. \textit{person}, \textit{car}, \textit{traffic light}, etc.), which include foreground elements that can  be copied onto other images without altering the plausibility of the scene, while excluding all the \textit{stuff} classes, which include background elements that cannot be easily moved across scenes. This categorization is similar to the one proposed in \cite{stuffAndThings} and we consider it easy to replicate in other datasets.

Then, we synthesize a new image $x^z$ and corresponding pseudo-labels $s^z$,  by assigning at each spatial location $p$ the candidate with the lowest depth, so that objects from different images will overlap plausibly into the synthesized one:
\begin{equation}
    x_{p}^z = x_{p}^k \quad s_{p}^z = s_{p}^k
\end{equation}
\begin{equation}
    k =
    \begin{cases}
        1, & D_{p} = \emptyset \\
        n \text{ s.t. } d_p^n = \min{D_{p}}, & D_{p} \neq \emptyset
    \end{cases}
\end{equation}
In Fig. \ref{fig:dbst} we depict our depth-based procedure to synthesize new training samples, considering, for the sake of simplicity, the case where $N$ is 2.

Hence, with the procedure detailed above, we synthesize an augmented version of \domainT{}, used to distill the whole \algoname{}-UDA framework into a single model by a self-training process. This dataset is much larger and exhibits more variability than the original \domainT{}. Due to its reliance on depth information, we dub our novel technique  as DBST (Depth-Based Self-Training). The results reported in Sec. \ref{sec:results} prove its remarkable effectiveness, both when used as the final stage following \algoname{} as well as when deployed as a standalone self-training procedure applied to any other UDA method.

\begin{table*}[t]
\begin{center}
\setlength{\tabcolsep}{2.5pt}
\scalebox{0.7}{
\begin{tabular}{l|ccccccccccccccccccc|cc}
\hline
Method &\rotatebox{90}{Road} & \rotatebox{90}{Sidewalk} & \rotatebox{90}{Building} &\rotatebox{90}{Walls} & \rotatebox{90}{Fence} & \rotatebox{90}{Pole} & \rotatebox{90}{T-light} & \rotatebox{90}{T-sign} & \rotatebox{90}{Vegetation} & \rotatebox{90}{Terrain} & \rotatebox{90}{Sky} & \rotatebox{90}{Person} 
& \rotatebox{90}{Rider} & \rotatebox{90}{Car}  & \rotatebox{90}{Truck} & \rotatebox{90}{Bus} & \rotatebox{90}{Train} & \rotatebox{90}{Motorbike} & \rotatebox{90}{Bicycle} & \textbf{mIoU} & \textbf{Acc} \\ \hline
AdaptSegNet~\cite{adaptsegnet} & 86.5 & 36.0 & 79.9 & 23.4 & 23.3 & 23.9 & 35.2 & 14.8 & 83.4 & 33.3 & 75.6 & 58.5 & 27.6 & 73.6 & 32.5 & 35.4 & 3.9 & 30.1 & 28.1 & 42.4 & 85.6 \\ 
\algoname{}-AdaptSegNet + DBST & 93.1 & 53.0 & 85.1 & 42.8 & 27.3 & 35.8 & 43.9 & 18.5 & 85.9 & 39.0 & 89.9 & 63.0 & 31.6 & 86.6 & 39.8 & 36.7 & 0 & 42.4 & 35.0 & \textbf{50.0} & \textbf{90.3} \\ \hline
MaxSquare~\cite{maxsquare} & 88.1 & 27.7 & 80.8 & 28.7 & 19.8 & 24.9 & 34.0 & 17.8 & 83.6 & 34.7 & 76.0 & 58.6 & 28.6 & 84.1 & 37.8 & 43.1 & 7.2 & 32.2 & 34.5 & 44.3 & 86.9 \\ 
\algoname{}-MaxSquare + DBST & 92.9 & 51.2 & 84.7 & 43.5 & 22.2 & 35.7 & 42.5 & 20.0 & 86.2 & 42.0 & 90.0 & 63.7 & 33.0 & 86.9 & 45.5 & 50.9 & 0 & 42.2 & 41.4 & \textbf{51.3}  & \textbf{90.3} \\ \hline
BDL~\cite{bdl} & 88.2 & 44.7 & 84.2 & 34.6 & 27.6 & 30.2 & 36.0 & 36.0 & 85.0 & 43.6 & 83.0 & 58.6 & 31.6 & 83.3 & 35.3 & 49.7 & 3.3 & 28.8 & 35.6 & 48.5 & 89.2 \\
\algoname{}-BDL + DBST & 93.2 & 52.6 & 86.4 & 44.1 & 31.2 & 36.5 & 42.4 & 36.1 & 86.3 & 41.0 & 89.8 & 63.3 & 37.4 & 86.3 & 42.8 & 57.8 & 0 & 40.3 & 37.9 & \textbf{52.9} & \textbf{90.7} \\ \hline
MRNET~\cite{mrnet} & 90.5 & 35.0 & 84.6 & 34.3 & 24.0 & 36.8 & 44.1 & 42.7 & 84.5 & 33.6 & 82.5 & 63.1 & 34.4 & 85.8 & 32.9 & 38.2 & 2.0 & 27.1 & 41.8 & 48.3  & 88.3        \\
\algoname{}-MRNET + DBST & 93.2 & 51.6 & 86.1 & 45.9 & 24.5 & 37.9 & 47.4 & 40.4 & 85.3 & 37.5 & 89.6 & 64.7 & 39.8 & 85.8 & 41.1 & 53.2 & 8.9 & 17.1 & 33.4 & \textbf{51.7}  & \textbf{90.0} \\ \hline
Stuff and things*~\cite{stuffAndThings} & 90.2 & 43.5 & 84.6 & 37.0 & 32.0 & 34.0 & 39.3 & 37.2 & 84.0 & 43.1 & 86.1 & 61.1 & 29.9 & 81.6 & 32.3 & 38.3 & 3.2  & 30.2 & 31.9 & 48.3 & 88.8 \\ 
\algoname{}-Stuff and things + DBST & 93.3 & 54.0 & 86.5 & 46.4 & 32.3 & 37.7 & 45.2 & 39.5 & 85.5 & 39.4 & 90.0 & 63.7 & 32.8 & 85.5 & 32.0 & 39.5 & 0 & 37.7 & 35.5 & \textbf{51.4} & \textbf{90.5} \\ \hline
FADA~\cite{fada} & 92.5 & 47.5 & 85.1 & 37.6 & 32.8 & 33.4 & 33.8 & 18.4 & 85.3 & 37.7 & 83.5 & 63.2 & 39.7 & 87.5 & 32.9 & 47.8 & 1.6 & 34.9 & 39.5 & 49.2 & 88.9 \\
\algoname{}-FADA + DBST & 93.9 & 58.2 & 86.4 & 45.9 & 29.6 & 36.9 & 44.6 & 27.0 & 86.3 & 39.4 & 90.0 & 64.9 & 41.0 & 85.8 & 34.6 & 51.2 & 9.9  & 24.2 & 37.3 & \textbf{52.0}  & \textbf{90.7} \\ \hline
LTIR~\cite{ltir} & 92.9 & 55.0 & 85.3 & 34.2 & 31.1 & 34.4 & 40.8 & 34.0 & 85.2 & 40.1 & 87.1 & 61.1 & 31.1 & 82.5 & 32.3 & 42.9 & 3 & 36.4 & 46.1 & 50.2 & 90.0 \\ 
\algoname{}-LTIR + DBST & 94.2 & 59.6 & 86.9 & 43.9 & 35.3 & 36.9 & 45.7 & 36.1 & 86.2 & 40.6 & 90.0 & 65.9 & 38.2 & 84.4 & 33.3 & 52.4 & 13.7 & 46.2 & 51.7 & \textbf{54.1}  & \textbf{91.0} \\ \hline
ProDA~\cite{zhang2021prototypical} & 87.8 & 56.0 & 79.7 & 46.3 & 44.8 & 45.6 & 53.5 & 53.5 & 88.6 & 45.2 & 82.1 & 70.7 & 39.2 & 88.8 & 45.5  & 59.4 & 1.0 & 48.9 & 56.4 & 57.5 & 89.1 \\ 
\algoname{}-ProDA + DBST & 94.3 & 60.0 & 87.9 & 50.5 & 43.0 & 42.6 & 50.8 & 51.3 & 88.0 & 45.9 & 89.7 & 68.9 & 41.8 & 88.0 & 45.8  & 63.8 & 0 & 50.0 & 55.8 & \textbf{58.8} & \textbf{92.1} \\
\hline
\end{tabular}}
\end{center}
\caption{Results on \gtacs{}. When available, checkpoints provided by authors are used. * denotes method retrained by us.}
\label{tab:results_gta}
\end{table*}

\begin{table*}[t]
\begin{center}
\setlength{\tabcolsep}{2.5pt}
\scalebox{0.75}{
\begin{tabular}{l|cccccccccccc|cc}
\hline
Method &\rotatebox{90}{Sky} & \rotatebox{90}{Building} & \rotatebox{90}{Road} &\rotatebox{90}{Sidewalk} & \rotatebox{90}{Fence} & \rotatebox{90}{Vegetation} & \rotatebox{90}{Pole} & \rotatebox{90}{Car} & \rotatebox{90}{T-Sign} & \rotatebox{90}{Person} & \rotatebox{90}{Bicycle} & \rotatebox{90}{T-Light}& \textbf{mIoU} & \textbf{Acc} \\
\hline
AdaptSegNet*~\cite{adaptsegnet} & 75.6 & 78.0 & 89.7 & 28.5 & 3.4 & 76.0 & 28.5 & 85.1 & 27.2 & 55.3 & 46.6 & 0 & 49.5 & 86.9 \\ 
\algoname{}-AdaptSegNet + DBST & 87.7 & 80.1 & 94.0 & 61.8 & 66.0 & 81.1 & 32.2 & 85.4 & 31.3 & 59.0 & 52.3 & 0   & \textbf{55.9} & \textbf{90.2} \\ \hline
MaxSquare*~\cite{maxsquare} & 72.4 & 79.2 & 89.2 & 36.0 & 4.6 & 75.7 & 31.5 & 84.9  & 30.7 & 55.8 & 45.8 & 8.6 & 51.2 & 87.3 \\
\algoname{}-MaxSquare + DBST & 87.5 & 80.0 & 93.7 & 61.8 & 7.3 & 80.8 & 33.2 & 84.6 & 35.1 & 58.1 & 48.1 & 8.2 & \textbf{56.5} & \textbf{90.1} \\ \hline
MRNET*~\cite{mrnet} & 84.6 & 79.7 & 93.9 & 56.3 & 0 & 80.5 & 35.4 & 88.9  & 27.2 & 59.4 & 56.3 & 0 & 54.5 & 90.0 \\
\algoname{}-MRNET + DBST & 88.3 & 79.9 & 93.9 & 63.0 & 6.3 & 81.3 & 35.5 & 84.3 & 31.3 & 59.5 & 47.9 & 0 & \textbf{55.9} & \textbf{90.2} \\ \hline
\end{tabular}}
\end{center}
\caption{Results on \synthiacs{}. * denotes method retrained by us.}
\label{tab:results_synthia}
\end{table*}

\section{Experiments}

\subsection{Implementation Details}
\label{sec:implementation}

\textbf{Network Architectures.}
We use Monodepth2~\cite{monodepth2} to generate depth proxy-labels for the procedure described in Sec. \ref{sec:d2s}.
We adapt the general framework presented in \cite{atdt} to our setting by deploying the popular Deeplab-v2~\cite{deeplabv2} for depth estimation and semantic segmentation networks. Both networks consist of a backbone and an ASPP module~\cite{deeplabv2}, which substitute, respectively, the encoder and decoder used in \cite{atdt}. The backbone is implemented as a dilated ResNet50~\cite{drn}.
We also remove the downsampling and upsampling operations used in \cite{atdt} when learning the transfer function between depth and semantics. More precisely, in our architecture the transfer function is realized as a simple 6-layers CNN with kernel size $3\times3$ and Batch Norm~\cite{batchnorm}. Following the recent trend in UDA for semantic segmentation \cite{adaptsegnet, maxsquare, bdl, mrnet, stuffAndThings, fada, ltir}, during DBST we train a single Deeplab-v2~\cite{deeplabv2} model, with a dilated ResNet101 pre-trained on Imagenet~\cite{imagenet} as backbone.

\textbf{Training Details.}
Our pipeline is implemented using PyTorch~\cite{pytorch} and trained on one NVIDIA Tesla V100 GPU with 16GB of memory. 
In every training and test phase we resize input images to 1024$\times$512, with the exception of DBST, when we first perform random scaling and then random crop with size 1024$\times$512. During DBST we use also color jitter to avoid overfitting on the pseudo-labels.
In our version of \cite{atdt}, the depth and the transfer network are optimized by Adam~\cite{adam} with batch size 2 for 70 and 40 epochs, respectively, while the semantic segmentation network is trained by SGD with batch size 2 for 70 epochs.The final model obtained by DBST is trained again with SGD, batch size 3 and for 30 epochs.
We adopt the One Cycle learning rate policy~\cite{onecycle} in every training, setting the maximum learning rate to $10^{-4}$ but in DBST, where we use $10^{-3}$.

\subsection{Datasets}
\label{sec:datasets}
We briefly describe the datasets adopted in our experiments, pointing to the Suppl. Mat. for additional details. We follow common practice \cite{adaptsegnet, ltir, bdl} and test our framework in the synthetic-to-real case using GTA5 \cite{gtabench, gta} or SYNTHIA \cite{synthia} as synthetic datasets. The former consists in synthetic images captured with the game Grand Theft Auto V, while the latter is composed of images generated by rendering a virtual city. Since our method requires video sequences to train Monodepth2~\cite{monodepth2}, we use the split SYNTHIA VIDEO SEQUENCES (SYNTHIA-SEQ) in the experiments involving the SYNTHIA dataset. As for real images, we leverage the popular Cityscapes dataset~\cite{Cityscapes}, which consists in a large collection of video sequences of driving scenes from 50 different cities in Germany. 
% All the mentioned datasets provide images labelled with fine semantic annotations, which we exploit for training (GTA5 or SYNTHIA-SEQ) and testing (Cityscapes).

\begin{figure*}[t]
    \centering
    \includegraphics[width=0.7\linewidth]{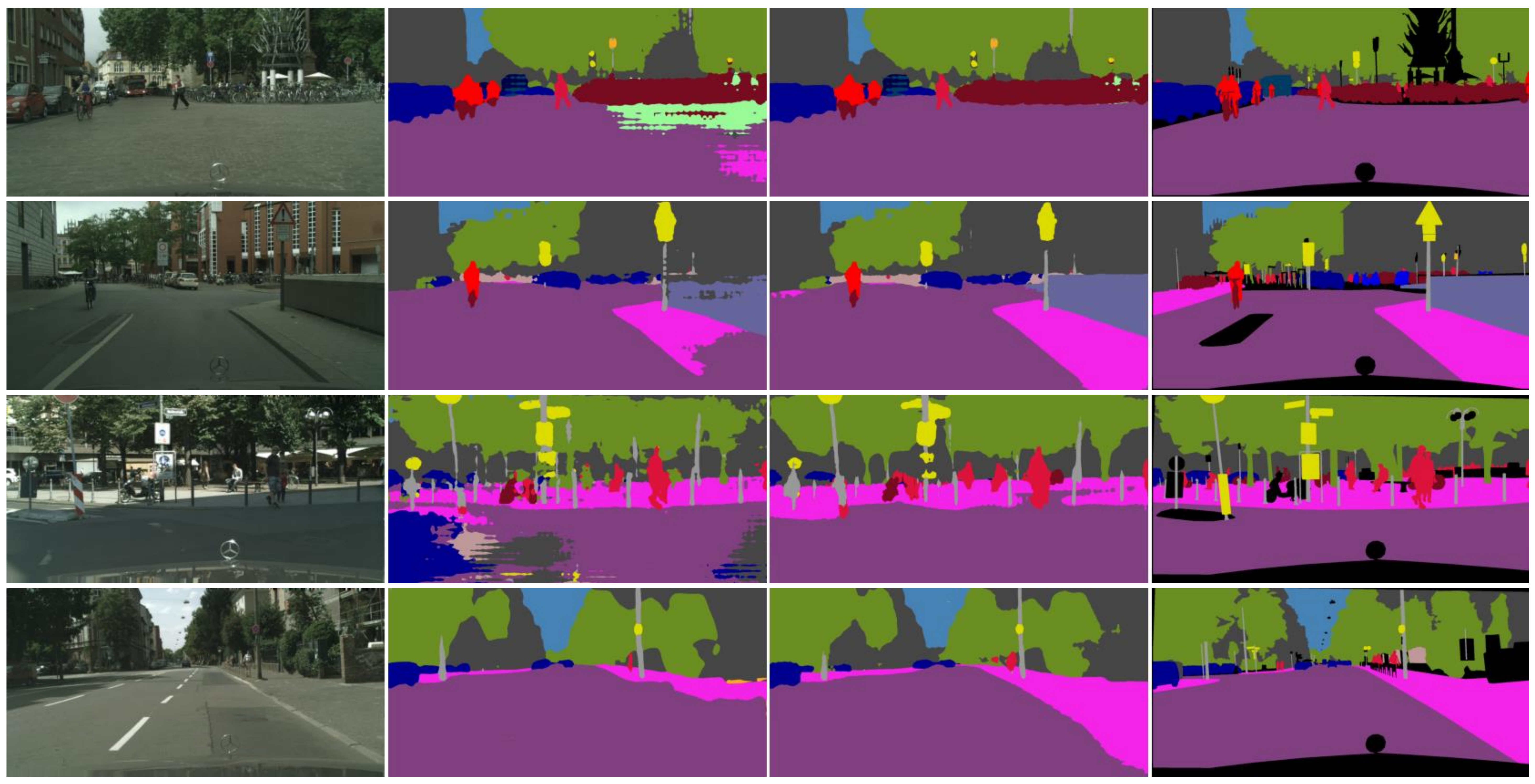}
    \caption{From left to right: RGB image, prediction from UDA method, prediction from \algoname{}-UDA + DBST, GT. The top two rows deal with \gtacs{}, the other two with  \synthiacs{}. Selected methods are, from top to bottom: LTIR~\cite{ltir}, BDL~\cite{bdl}, MaxSquare~\cite{maxsquare} and MRNET~\cite{mrnet}. In all these examples our proposal  can ameliorate dramatically the output of the given stand-alone method, especially on classes featuring large and regular shapes, like \textit{road} in rows 1-3, \textit{sidewalk} in rows 2-4 and \textit{wall} in row 2.}
    \label{fig:qualitative}
\end{figure*}

\subsection{Results}

\label{sec:results}

We report here experimental results obtained in two domain adaptation benchmarks, which show how the combination with our \algoname{} method allows to boost performance of recent UDA for semantic segmentation approaches. 

\textbf{\gtacs{}.}
Tab. \ref{tab:results_gta} reports results on the most popular UDA benchmark for semantic segmentation, i.e. \gtacs{}, where methods are trained on GTA5 and tested on Cityscapes. We selected the most relevant UDA approaches proposed in the last years \cite{adaptsegnet, maxsquare, bdl, mrnet, stuffAndThings, fada, ltir, zhang2021prototypical}, using checkpoints provided by authors when available. We report per-class and overall results in terms of mean intersection over union (mIoU) and pixel accuracy (Acc), when each method is either used stand-alone or deployed  within our proposal (i.e. \algoname{} + DBST). The reader may notice how every UDA method does improve considerably if combined with our proposal, despite the variability of their stand-alone performances. Indeed, AdaptSegNet~\cite{adaptsegnet}, which yields about 42 in terms of mIoU, reaches 50 when embedded into our framework. Likewise, ProDA, currently considered the s.o.t.a. UDA method, improves in  mIoU from 57.5 to 58.8. Moreover, we can observe in Tab. \ref{tab:results_gta} that our method produces a general improvement for all classes, although we experience a certain performance variability for some of them (such as \textit{train}, \textit{motorbike} and \textit{bicycle}), probably due to noisy pseudo-labels used during DBST. Conversely, our method yields consistently a significant gain on classes characterized by large and regular shapes, namely \textit{road}, \textit{sidewalk}, \textit{building}, \textit{wall} and \textit{sky}. This validates the effectiveness of a) the  geometric cues derivable from depth to predict the semantics of these kind of objects and b) the methodology we propose to leverage  on these additional cues in UDA settings. This behavior is also clearly observable from qualitatives in Fig. \ref{fig:qualitative}. We point out that, to the best of our knowledge, the performance obtained by \algoname{}-ProDA + DBST,  i.e. 58.8 mIoU (last row of Tab. \ref{tab:results_gta}) establishes the new state-of-the-art for \gtacs{}. 

\textbf{\synthiacs{}.}
Akin to common practice in literature we present results also  on the popular SYNTHIA dataset. As our pipeline requires video sequences to train the self-supervised monocular depth estimation network, we select the SYNTHIA VIDEO SEQUENCES split for training and the Cityscapes dataset for testing. We will call this setting \synthiacs{}. To address it, we re-trained the UDA methods for which the code is available and the training procedure is more affordable in terms of memory and run-time requirements,  namely AdaptSegNet~\cite{adaptsegnet}, MaxSquare~\cite{maxsquare} and MRNET~\cite{mrnet}. The results in Tab. \ref{tab:results_synthia} show that all the selected UDA approaches exhibit a substantial performance gain when coupled with our proposal, with a general improvement in all classes. In particular, similarly to the results obtained in \gtacs{}, we observe a consistent improvement for classes related to objects with large and regular shapes (as depicted also  in Fig. \ref{fig:qualitative}), with the only exception of a slight performance drop for the class \textit{building} when using MRNET~\cite{mrnet} (last row of Tab. \ref{tab:results_synthia}). We argue that our approach is relatively less effective with  MRNET~\cite{mrnet} as, unlike AdaptSegNet~\cite{adaptsegnet} and MaxSquare~\cite{maxsquare}, it  yields  already satisfactory results in those classes  which are usually improved by the geometric clues injected by \algoname{}.

In the Suppl. Mat. we show that it is also possible to exploit the depth ground-truths provided by the SYNTHIA dataset as an additional source of supervision during the training of Monodepth2~\cite{monodepth2}, obtaining a small improvement in the performances of the overall framework.

\subsection{Analysis}
\label{sec:Analysis}
We report here the most relevant analysis concerning our work. Additional ones can be found in the Suppl. Mat..

\textbf{Ablation studies.}
In Tab. \ref{tab:ablation1_gta}, we analyze the impact on the performance of our two main contributions, i.e. injection of geometric cues into UDA methods by \algoname{} and DBST. Purposely, we select the \gtacs{} benchmark and, for the top performing UDA methods, we report the mIoU figures obtained by using the stand-alone UDA method (column \textit{UDA}), combining it with \algoname{} (column \textit{D4-UDA}), applying DBST directly on the stand-alone method (column \textit{UDA + DBST}) and embedding the method into our full pipeline (column \textit{D4-UDA + DBST}). We can observe that each of our novel contributions improves the performance of the most recent UDA methods by a large margin, which is even more remarkable considering that the selected methods already include one or more step of self-training. Moreover, \algoname{} and DBST further enhance the performances of any selected method when deployed jointly, as shown in the column \textit{D4-UDA + DBST}, suggesting that they are complementary.
In order to further assess the effectiveness of DBST, in the column \textit{D4-UDA + ST} we report results obtained by D4-UDA in combination with a baseline self-training procedure, which consists in simply fine-tuning the model by its own predictions on the images of the target domain. As the only difference between this procedure and our DBST is the dataset employed for fine-tuning, the results prove the effectiveness of DBST in generating a varied set of plausible samples more amenable to self-training than the original images belonging to the target domain.

\textbf{Alternative strategies to exploit depth.} As explained in Sec. \ref{sec:d2s} \textit{Semantics from Depth}, we rely on the mechanism of transferring features across tasks and domains from \cite{atdt} to inject depth cues into semantic segmentation. To validate our choice, we explore two possible alternatives, namely DeepLabV2-RGBD and DeepLabV2-Depth. Both consist in the popular DeepLabV2~\cite{deeplabv2} network, with RGBD images in input in the first case and depth maps (no RGB) in the second (more details in the Suppl. Mat.). Tab. \ref{tab:atdt_alternatives} compares the performance of these alternatives with our method, either when used standalone (rows 2, 3, and 4) or when combined with LTIR~\cite{ltir} according to the strategy presented in Sec. \ref{sec:d2s} \textit{Combine with UDA}.
Results allow us to make some important considerations. First, our intuition on the possibility of exploiting depth to improve semantics is correct since also simple approaches improve over the baseline (reported in the first row of the table). Nonetheless, these naive methods produce a significantly smaller improvement compared to our approach, showing that our decision to adapt \cite{atdt} to the UDA scenario is not obvious. Moreover, \cite{atdt} requires only RGB images at test time. Finally, when combined with LTIR~\cite{ltir}, a stronger depth-to-semantic model provides better results, validating our choice once again.

\newcolumntype{C}[1]{>{\centering\arraybackslash}p{#1}}

\begin{table}
\begin{center}
\scalebox{0.65}{
\begin{tabular}{c |C{1cm} C{1.4cm} C{1.1cm} C{1.6cm} | C{1.6cm}}
\hline
Method & UDA & \algoname{}-UDA & UDA + DBST & \algoname{}-UDA + DBST & \algoname{}-UDA + ST\\
\hline
BDL~\cite{bdl} & 48.5 & 49.6 & 51.7 & \textbf{52.9} & 50.1\\
MRNET~\cite{mrnet} & 48.3 & 49.6 & 50.0 &\textbf{51.7} & 50.3\\
Stuff and Things*~\cite{stuffAndThings} & 48.3 & 49.1 & 50.4 & \textbf{51.4} & 49.4\\
FADA~\cite{fada}  & 49.3 & 49.9 & 51.4 & \textbf{52.0} & 50.0\\
LTIR~\cite{ltir}  & 50.2 & 51.1 & 53.1 &\textbf{54.1} & 51.5\\
ProDa~\cite{zhang2021prototypical} & 57.5 & 57.6 & 58.0 & \textbf{58.8} & 56.8\\
\hline 
\end{tabular}}
\end{center}
\caption{Impact on performance of the two components of our proposal (\algoname{}, DBST) when applied separately or jointly to selected UDA methods on \gtacs{}. * indicates that the method was retrained by us. Results are reported in terms of mIoU.} 
\label{tab:ablation1_gta}
\end{table}

\begin{table}
\begin{center}
\scalebox{0.6}{
\begin{tabular}{l | c}
\hline
Method & mIoU \\
\hline
DeepLabV2-RGB & 34.5\\
\hline
DeepLabV2-RGBD & 35.5 \\
DeepLabV2-Depth & 36.5 \\
Semantics from depth (Sec. \ref{sec:d2s}) & \textbf{43.1} \\
\hline                                    
DeepLabV2-RGBD $\bigoplus{}$ LTIR \cite{ltir} & 47.7\\
DeepLabV2-Depth $\bigoplus{}$ LTIR \cite{ltir} & 49.3\\
D4-LTIR & \textbf{51.1} \\
\hline
\end{tabular}}
\end{center}
\caption{Comparison between alternative methods to infer semantics from depth. DeepLabV2-RGB, DeepLabV2-RGBD and DeepLabV2-Depth stand for DeepLabV2 \cite{deeplabv2} trained on \domainS{}, using respectively RGB images, RGBD images or depth proxy-labels as input, while ``Semantics from depth'' is the approach described in Sec. \ref{sec:d2s} \textit{Semantics from Depth}. The symbol $\bigoplus{}$ represents the merge operation described in Sec. \ref{sec:d2s} \textit{Combine with UDA}. Results are reported in terms of mIoU on the Cityscapes dataset.}
\label{tab:atdt_alternatives}
\end{table}

\textbf{Impact of video sequences.} As described in Sec. \ref{sec:d2s}, we obtain depth proxy-labels with a self-supervised depth estimation network \cite{monodepth2}, that we train using the raw video sequences (just RGB images) provided by the datasets involved in our experiments. In order to validate that using video sequences from the target domain doesn't provide any advantage to our framework, we train AdaptSegNet~\cite{adaptsegnet} on \gtacs{} using the whole training split available for Cityscapes (i.e. 83300 images with temporal consistency). We choose AdaptSegNet~\cite{adaptsegnet} since it can be considered the building block of many UDA methods. We observe a drop in performances from 42.4 to 41.9 mIoU, showing that using video sequences does not boost semantic segmentation in a UDA setting, probably because of the similarity between consecutive frames, and that the improvement produced by our framework is provided by the effective strategy that we adopt to exploit depth.

\section{Conclusion}
We have shown how to exploit self-supervised monocular depth estimation in UDA problems to obtain accurate semantic predictions for objects with strong geometric priors (like road and buildings). As all recent UDA approaches lack such geometric knowledge, we build our \algoname{} method as a depth-based add-on, pluggable into any UDA method to boost performances. Finally, we employed self-supervised depth estimation to realize an effective data augmentation strategy for self-training. Our work highlights the possibility of exploiting auxiliary tasks learned by self-supervision to better tackle UDA for semantic segmentation, paving the way for novel research directions. 

{\small
\bibliographystyle{ieee_fullname}
\bibliography{egbib}

\begin{thebibliography}{10}\itemsep=-1pt

\bibitem{michieli}
Matteo Biasetton, Umberto Michieli, Gianluca Agresti, and Pietro Zanuttigh.
\newblock Unsupervised domain adaptation for semantic segmentation of urban
  scenes.
\newblock In {\em Proceedings of the IEEE/CVF Conference on Computer Vision and
  Pattern Recognition (CVPR) Workshops}, June 2019.

\bibitem{Chang_2019}
Wei-Lun Chang, Hui-Po Wang, Wen-Hsiao Peng, and Wei-Chen Chiu.
\newblock All about structure: Adapting structural information across domains
  for boosting semantic segmentation.
\newblock {\em 2019 IEEE/CVF Conference on Computer Vision and Pattern
  Recognition (CVPR)}, Jun 2019.

\bibitem{deeplabv2}
Liang-Chieh Chen, George Papandreou, Iasonas Kokkinos, Kevin Murphy, and
  Alan~L. Yuille.
\newblock Deeplab: Semantic image segmentation with deep convolutional nets,
  atrous convolution, and fully connected crfs.
\newblock {\em IEEE Transactions on Pattern Analysis and Machine Intelligence},
  40(4):834–848, Apr 2018.

\bibitem{deeplabv3+}
Liang-Chieh Chen, Yukun Zhu, George Papandreou, Florian Schroff, and Hartwig
  Adam.
\newblock Encoder-decoder with atrous separable convolution for semantic image
  segmentation.
\newblock In {\em Proceedings of the European conference on computer vision
  (ECCV)}, pages 801--818, 2018.

\bibitem{maxsquare}
Minghao Chen, Hongyang Xue, and Deng Cai.
\newblock Domain adaptation for semantic segmentation with maximum squares
  loss.
\newblock {\em 2019 IEEE/CVF International Conference on Computer Vision
  (ICCV)}, Oct 2019.

\bibitem{geoguided}
Yuhua Chen, Wen Li, Xiaoran Chen, and Luc~Van Gool.
\newblock Learning semantic segmentation from synthetic data: A geometrically
  guided input-output adaptation approach.
\newblock In {\em Proceedings of the IEEE Conference on Computer Vision and
  Pattern Recognition}, pages 1841--1850, 2019.

\bibitem{inout_adaptation}
Yuhua Chen, Wen Li, Xiaoran Chen, and Luc Van~Gool.
\newblock Learning semantic segmentation from synthetic data: A geometrically
  guided input-output adaptation approach.
\newblock {\em 2019 IEEE/CVF Conference on Computer Vision and Pattern
  Recognition (CVPR)}, Jun 2019.

\bibitem{road}
Yuhua Chen, Wen Li, and Luc~Van Gool.
\newblock Road: Reality oriented adaptation for semantic segmentation of urban
  scenes.
\newblock {\em 2018 IEEE/CVF Conference on Computer Vision and Pattern
  Recognition}, Jun 2018.

\bibitem{Choi_2019}
Jaehoon Choi, Taekyung Kim, and Changick Kim.
\newblock Self-ensembling with gan-based data augmentation for domain
  adaptation in semantic segmentation.
\newblock {\em 2019 IEEE/CVF International Conference on Computer Vision
  (ICCV)}, Oct 2019.

\bibitem{Cityscapes}
Marius Cordts, Mohamed Omran, Sebastian Ramos, Timo Rehfeld, Markus Enzweiler,
  Rodrigo Benenson, Uwe Franke, Stefan Roth, and Bernt Schiele.
\newblock The cityscapes dataset for semantic urban scene understanding.
\newblock In {\em The IEEE Conference on Computer Vision and Pattern
  Recognition (CVPR)}, June 2016.

\bibitem{imagenet}
J. {Deng}, W. {Dong}, R. {Socher}, L. {Li}, {Kai Li}, and {Li Fei-Fei}.
\newblock Imagenet: A large-scale hierarchical image database.
\newblock In {\em 2009 IEEE Conference on Computer Vision and Pattern
  Recognition}, pages 248--255, 2009.

\bibitem{Garg_2016}
Ravi Garg, Vijay~Kumar B.G., Gustavo Carneiro, and Ian Reid.
\newblock Unsupervised cnn for single view depth estimation: Geometry to the
  rescue.
\newblock {\em Lecture Notes in Computer Science}, page 740–756, 2016.

\bibitem{monodepth}
Clement Godard, Oisin~Mac Aodha, and Gabriel~J. Brostow.
\newblock Unsupervised monocular depth estimation with left-right consistency.
\newblock {\em 2017 IEEE Conference on Computer Vision and Pattern Recognition
  (CVPR)}, Jul 2017.

\bibitem{monodepth2}
Clement Godard, Oisin~Mac Aodha, Michael Firman, and Gabriel Brostow.
\newblock Digging into self-supervised monocular depth estimation.
\newblock {\em 2019 IEEE/CVF International Conference on Computer Vision
  (ICCV)}, Oct 2019.

\bibitem{semanticallyguided}
Vitor Guizilini, Rui Hou, Jie Li, Rares Ambrus, and Adrien Gaidon.
\newblock Semantically-guided representation learning for self-supervised
  monocular depth.
\newblock In {\em Proceedings of the Eighth International Conference on
  Learning Representations (ICLR)}, 2020.

\bibitem{resnet}
K. {He}, X. {Zhang}, S. {Ren}, and J. {Sun}.
\newblock Deep residual learning for image recognition.
\newblock In {\em 2016 IEEE Conference on Computer Vision and Pattern
  Recognition (CVPR)}, pages 770--778, 2016.

\bibitem{cycada}
Judy Hoffman, E. Tzeng, T. Park, Jun-Yan Zhu, Phillip Isola, Kate Saenko,
  Alexei~A. Efros, and Trevor Darrell.
\newblock Cycada: Cycle-consistent adversarial domain adaptation.
\newblock In {\em ICML}, 2018.

\bibitem{hoffman2016fcns}
Judy Hoffman, Dequan Wang, Fisher Yu, and Trevor Darrell.
\newblock Fcns in the wild: Pixel-level adversarial and constraint-based
  adaptation, 2016.

\bibitem{hoyer2021three}
Lukas Hoyer, Dengxin Dai, Yuhua Chen, Adrian Koring, Suman Saha, and Luc
  Van~Gool.
\newblock Three ways to improve semantic segmentation with self-supervised
  depth estimation.
\newblock In {\em Proceedings of the IEEE/CVF Conference on Computer Vision and
  Pattern Recognition}, pages 11130--11140, 2021.

\bibitem{batchnorm}
Sergey Ioffe and Christian Szegedy.
\newblock Batch normalization: Accelerating deep network training by reducing
  internal covariate shift, 2015.

\bibitem{semantic_booster}
Jianbo Jiao, Ying Cao, Yibing Song, and Rynson Lau.
\newblock Look deeper into depth: Monocular depth estimation with semantic
  booster and attention-driven loss.
\newblock In {\em Proceedings of the European Conference on Computer Vision
  (ECCV)}, September 2018.

\bibitem{ltir}
Myeongjin Kim and Hyeran Byun.
\newblock Learning texture invariant representation for domain adaptation of
  semantic segmentation.
\newblock {\em 2020 IEEE/CVF Conference on Computer Vision and Pattern
  Recognition (CVPR)}, Jun 2020.

\bibitem{adam}
Diederik~P. Kingma and Jimmy Ba.
\newblock Adam: A method for stochastic optimization.
\newblock In {\em 3rd International Conference on Learning Representations
  (ICLR)}, 2015.

\bibitem{Semantic_Guidance}
Marvin Klingner, Jan-Aike Term{\"o}hlen, Jonas Mikolajczyk, and Tim
  Fingscheidt.
\newblock Self-supervised monocular depth estimation: Solving the dynamic
  object problem by semantic guidance.
\newblock In {\em European Conference on Computer Vision}, pages 582--600.
  Springer, 2020.

\bibitem{kundu2019adapt}
Jogendra~Nath Kundu, Nishank Lakkakula, and R~Venkatesh Babu.
\newblock Um-adapt: Unsupervised multi-task adaptation using adversarial
  cross-task distillation.
\newblock In {\em Proceedings of the IEEE International Conference on Computer
  Vision}, pages 1436--1445, 2019.

\bibitem{pseudolabel}
D. Lee.
\newblock Pseudo-label : The simple and efficient semi-supervised learning
  method for deep neural networks.
\newblock In {\em Workshop on challenges in representation learning, ICML},
  2013.

\bibitem{spigan}
Kuan-Hui Lee, German Ros, Jie Li, and Adrien Gaidon.
\newblock Spigan: Privileged adversarial learning from simulation.
\newblock In {\em International Conference on Learning Representations}, 2019.

\bibitem{bdl}
Yunsheng Li, Lu Yuan, and Nuno Vasconcelos.
\newblock Bidirectional learning for domain adaptation of semantic
  segmentation.
\newblock {\em 2019 IEEE/CVF Conference on Computer Vision and Pattern
  Recognition (CVPR)}, Jun 2019.

\bibitem{pycda}
Qing Lian, Lixin Duan, Fengmao Lv, and Boqing Gong.
\newblock Constructing self-motivated pyramid curriculums for cross-domain
  semantic segmentation: A non-adversarial approach.
\newblock {\em 2019 IEEE/CVF International Conference on Computer Vision
  (ICCV)}, Oct 2019.

\bibitem{instance_adaptive}
Ke Mei, Chuang Zhu, Jiaqi Zou, and Shanghang Zhang.
\newblock Instance adaptive self-training for unsupervised domain adaptation.
\newblock In {\em The European Conference on Computer Vision (ECCV)}, August
  2020.

\bibitem{Murez_2018}
Zak Murez, Soheil Kolouri, David Kriegman, Ravi Ramamoorthi, and Kyungnam Kim.
\newblock Image to image translation for domain adaptation.
\newblock {\em 2018 IEEE/CVF Conference on Computer Vision and Pattern
  Recognition}, Jun 2018.

\bibitem{classmix}
Viktor Olsson, Wilhelm Tranheden, Juliano Pinto, and Lennart Svensson.
\newblock Classmix: Segmentation-based data augmentation for semi-supervised
  learning, 2020.

\bibitem{Pan_2020}
Fei Pan, Inkyu Shin, Francois Rameau, Seokju Lee, and In~So Kweon.
\newblock Unsupervised intra-domain adaptation for semantic segmentation
  through self-supervision.
\newblock {\em 2020 IEEE/CVF Conference on Computer Vision and Pattern
  Recognition (CVPR)}, Jun 2020.

\bibitem{Enet}
Adam Paszke, Abhishek Chaurasia, Sangpil Kim, and Eugenio Culurciello.
\newblock Enet: {A} deep neural network architecture for real-time semantic
  segmentation.
\newblock {\em CoRR}, abs/1606.02147, 2016.

\bibitem{pytorch}
Adam Paszke, Sam Gross, Soumith Chintala, Gregory Chanan, Edward Yang, Zachary
  DeVito, Zeming Lin, Alban Desmaison, Luca Antiga, and Adam Lerer.
\newblock Automatic differentiation in pytorch.
\newblock In {\em NIPS 2017 Workshop on Autodiff}, 2017.

\bibitem{pizzati}
Fabio Pizzati, Raoul~de Charette, Michela Zaccaria, and Pietro Cerri.
\newblock Domain bridge for unpaired image-to-image translation and
  unsupervised domain adaptation.
\newblock In {\em Proceedings of the IEEE/CVF Winter Conference on Applications
  of Computer Vision (WACV)}, March 2020.

\bibitem{geometry}
Pierluigi~Zama Ramirez, Matteo Poggi, Fabio Tosi, Stefano Mattoccia, and Luigi
  Di~Stefano.
\newblock Geometry meets semantics for semi-supervised monocular depth
  estimation.
\newblock In {\em Asian Conference on Computer Vision}, pages 298--313.
  Springer, 2018.

\bibitem{atdt}
Pierluigi~Zama Ramirez, Alessio Tonioni, Samuele Salti, and Luigi~Di Stefano.
\newblock Learning across tasks and domains.
\newblock {\em 2019 IEEE/CVF International Conference on Computer Vision
  (ICCV)}, Oct 2019.

\bibitem{gtabench}
Stephan~R. Richter, Zeeshan Hayder, and Vladlen Koltun.
\newblock Playing for benchmarks.
\newblock In {\em {IEEE} International Conference on Computer Vision, {ICCV}
  2017, Venice, Italy, October 22-29, 2017}, pages 2232--2241, 2017.

\bibitem{gta}
Stephan~R. Richter, Vibhav Vineet, Stefan Roth, and Vladlen Koltun.
\newblock Playing for data: Ground truth from computer games.
\newblock {\em Lecture Notes in Computer Science}, page 102–118, 2016.

\bibitem{unet}
Olaf Ronneberger, Philipp Fischer, and Thomas Brox.
\newblock U-net: Convolutional networks for biomedical image segmentation.
\newblock {\em Medical Image Computing and Computer-Assisted Intervention –
  MICCAI 2015}, page 234–241, 2015.

\bibitem{synthia}
German Ros, Laura Sellart, Joanna Materzynska, David Vazquez, and Antonio~M.
  Lopez.
\newblock The synthia dataset: A large collection of synthetic images for
  semantic segmentation of urban scenes.
\newblock In {\em The IEEE Conference on Computer Vision and Pattern
  Recognition (CVPR)}, June 2016.

\bibitem{saha2021learning}
Suman Saha, Anton Obukhov, Danda~Pani Paudel, Menelaos Kanakis, Yuhua Chen,
  Stamatios Georgoulis, and Luc Van~Gool.
\newblock Learning to relate depth and semantics for unsupervised domain
  adaptation.
\newblock In {\em Proceedings of the IEEE/CVF Conference on Computer Vision and
  Pattern Recognition}, pages 8197--8207, 2021.

\bibitem{cnnss}
Evan Shelhamer, Jonathan Long, and Trevor Darrell.
\newblock Fully convolutional networks for semantic segmentation.
\newblock {\em IEEE Transactions on Pattern Analysis and Machine Intelligence},
  39(4):640–651, Apr 2017.

\bibitem{onecycle}
Leslie~N. Smith and Nicholay Topin.
\newblock Super-convergence: very fast training of neural networks using large
  learning rates.
\newblock {\em Artificial Intelligence and Machine Learning for Multi-Domain
  Operations Applications}, May 2019.

\bibitem{efficientdet}
Mingxing Tan, Ruoming Pang, and Quoc~V Le.
\newblock Efficientdet: Scalable and efficient object detection.
\newblock In {\em Proceedings of the IEEE/CVF Conference on Computer Vision and
  Pattern Recognition}, pages 10781--10790, 2020.

\bibitem{monoresmatch}
Fabio Tosi, Filippo Aleotti, Matteo Poggi, and Stefano Mattoccia.
\newblock Learning monocular depth estimation infusing traditional stereo
  knowledge.
\newblock In {\em Proceedings of the IEEE Conference on Computer Vision and
  Pattern Recognition}, pages 9799--9809, 2019.

\bibitem{dacs}
Wilhelm Tranheden, Viktor Olsson, Juliano Pinto, and Lennart Svensson.
\newblock Dacs: Domain adaptation via cross-domain mixed sampling.
\newblock In {\em Proceedings of the IEEE/CVF Winter Conference on Applications
  of Computer Vision (WACV)}, pages 1379--1389, January 2021.

\bibitem{adaptsegnet}
Yi-Hsuan Tsai, Wei-Chih Hung, Samuel Schulter, Kihyuk Sohn, Ming-Hsuan Yang,
  and Manmohan Chandraker.
\newblock Learning to adapt structured output space for semantic segmentation.
\newblock {\em 2018 IEEE/CVF Conference on Computer Vision and Pattern
  Recognition}, Jun 2018.

\bibitem{patch}
Yi-Hsuan Tsai, Kihyuk Sohn, Samuel Schulter, and Manmohan Chandraker.
\newblock Domain adaptation for structured output via discriminative patch
  representations.
\newblock {\em 2019 IEEE/CVF International Conference on Computer Vision
  (ICCV)}, Oct 2019.

\bibitem{tzeng2015simultaneous}
Eric Tzeng, Judy Hoffman, Trevor Darrell, and Kate Saenko.
\newblock Simultaneous deep transfer across domains and tasks.
\newblock In {\em Proceedings of the IEEE International Conference on Computer
  Vision}, pages 4068--4076, 2015.

\bibitem{ADVENT}
Tuan-Hung Vu, Himalaya Jain, Maxime Bucher, Matthieu Cord, and Patrick Perez.
\newblock Advent: Adversarial entropy minimization for domain adaptation in
  semantic segmentation.
\newblock {\em 2019 IEEE/CVF Conference on Computer Vision and Pattern
  Recognition (CVPR)}, Jun 2019.

\bibitem{dada}
Tuan-Hung Vu, Himalaya Jain, Maxime Bucher, Matthieu Cord, and Patrick~Perez
  Perez.
\newblock Dada: Depth-aware domain adaptation in semantic segmentation.
\newblock {\em 2019 IEEE/CVF International Conference on Computer Vision
  (ICCV)}, Oct 2019.

\bibitem{fada}
Haoran Wang, Tong Shen, Wei Zhang, Lingyu Duan, and Tao Mei.
\newblock Classes matter: A fine-grained adversarial approach to cross-domain
  semantic segmentation.
\newblock In {\em The European Conference on Computer Vision (ECCV)}, August
  2020.

\bibitem{stuffAndThings}
Zhonghao Wang, Mo Yu, Yunchao Wei, Rogerio Feris, Jinjun Xiong, Wen-mei Hwu,
  Thomas~S. Huang, and Honghui Shi.
\newblock Differential treatment for stuff and things: A simple unsupervised
  domain adaptation method for semantic segmentation.
\newblock {\em 2020 IEEE/CVF Conference on Computer Vision and Pattern
  Recognition (CVPR)}, Jun 2020.

\bibitem{multichannel}
Kohei Watanabe, Kuniaki Saito, Yoshitaka Ushiku, and Tatsuya Harada.
\newblock Multichannel semantic segmentation with unsupervised domain
  adaptation.
\newblock In {\em Proceedings of the European Conference on Computer Vision
  (ECCV)}, pages 0--0, 2018.

\bibitem{depthhints}
Jamie Watson, Michael Firman, Gabriel~J Brostow, and Daniyar Turmukhambetov.
\newblock Self-supervised monocular depth hints.
\newblock In {\em Proceedings of the IEEE International Conference on Computer
  Vision}, pages 2162--2171, 2019.

\bibitem{dcan}
Zuxuan Wu, Xintong Han, Yen-Liang Lin, Mustafa~Gökhan Uzunbas, Tom Goldstein,
  Ser~Nam Lim, and Larry~S. Davis.
\newblock Dcan: Dual channel-wise alignment networks for unsupervised scene
  adaptation.
\newblock {\em Lecture Notes in Computer Science}, page 535–552, 2018.

\bibitem{adversarial_perturbation}
Jihan Yang, Ruijia Xu, Ruiyu Li, Xiaojuan Qi, Xiaoyong Shen, Guanbin Li, and
  Liang Lin.
\newblock An adversarial perturbation oriented domain adaptation approach for
  semantic segmentation.
\newblock {\em Proceedings of the AAAI Conference on Artificial Intelligence},
  34(07):12613–12620, Apr 2020.

\bibitem{transferable}
Jason Yosinski, Jeff Clune, Yoshua Bengio, and Hod Lipson.
\newblock How transferable are features in deep neural networks?
\newblock In {\em Advances in neural information processing systems}, pages
  3320--3328, 2014.

\bibitem{drn}
Fisher Yu, Vladlen Koltun, and Thomas Funkhouser.
\newblock Dilated residual networks.
\newblock {\em 2017 IEEE Conference on Computer Vision and Pattern Recognition
  (CVPR)}, Jul 2017.

\bibitem{ipas}
Pierluigi Zama~Ramirez, Alessio Tonioni, and Luigi Di~Stefano.
\newblock Exploiting semantics in adversarial training for image-level domain
  adaptation.
\newblock In {\em 2018 IEEE International Conference on Image Processing,
  Applications and Systems (IPAS)}, pages 49--54, 2018.

\bibitem{taskonomy}
Amir~R Zamir, Alexander Sax, William Shen, Leonidas~J Guibas, Jitendra Malik,
  and Silvio Savarese.
\newblock Taskonomy: Disentangling task transfer learning.
\newblock In {\em Proceedings of the IEEE conference on computer vision and
  pattern recognition}, pages 3712--3722, 2018.

\bibitem{zhang2021prototypical}
Pan Zhang, Bo Zhang, Ting Zhang, Dong Chen, Yong Wang, and Fang Wen.
\newblock Prototypical pseudo label denoising and target structure learning for
  domain adaptive semantic segmentation.
\newblock {\em arXiv preprint arXiv:2101.10979}, 2021.

\bibitem{Zhang_2017}
Yang Zhang, Philip David, and Boqing Gong.
\newblock Curriculum domain adaptation for semantic segmentation of urban
  scenes.
\newblock {\em 2017 IEEE International Conference on Computer Vision (ICCV)},
  Oct 2017.

\bibitem{Zhang_2018}
Yiheng Zhang, Zhaofan Qiu, Ting Yao, Dong Liu, and Tao Mei.
\newblock Fully convolutional adaptation networks for semantic segmentation.
\newblock {\em 2018 IEEE/CVF Conference on Computer Vision and Pattern
  Recognition}, Jun 2018.

\bibitem{joint_adversarial}
Y. Zhang and Zilei Wang.
\newblock Joint adversarial learning for domain adaptation in semantic
  segmentation.
\newblock In {\em AAAI}, 2020.

\bibitem{rectifying}
Zhedong Zheng and Yi Yang.
\newblock Rectifying pseudo label learning via uncertainty estimation for
  domain adaptive semantic segmentation.
\newblock {\em International Journal of Computer Vision (IJCV)}, 2020.

\bibitem{mrnet}
Zhedong Zheng and Yi Yang.
\newblock Unsupervised scene adaptation with memory regularization in vivo.
\newblock {\em Proceedings of the Twenty-Ninth International Joint Conference
  on Artificial Intelligence}, Jul 2020.

\bibitem{Zhou_2017}
Tinghui Zhou, Matthew Brown, Noah Snavely, and David~G. Lowe.
\newblock Unsupervised learning of depth and ego-motion from video.
\newblock {\em 2017 IEEE Conference on Computer Vision and Pattern Recognition
  (CVPR)}, Jul 2017.

\bibitem{cyclegan}
Jun-Yan Zhu, Taesung Park, Phillip Isola, and Alexei~A. Efros.
\newblock Unpaired image-to-image translation using cycle-consistent
  adversarial networks.
\newblock {\em 2017 IEEE International Conference on Computer Vision (ICCV)},
  Oct 2017.

\bibitem{cbst}
Yang Zou, Zhiding Yu, BVK~Vijaya Kumar, and Jinsong Wang.
\newblock Unsupervised domain adaptation for semantic segmentation via
  class-balanced self-training.
\newblock In {\em Proceedings of the European Conference on Computer Vision
  (ECCV)}, pages 289--305, 2018.

\bibitem{crst}
Yang Zou, Zhiding Yu, Xiaofeng Liu, B.V.K.~Vijaya Kumar, and Jinsong Wang.
\newblock Confidence regularized self-training.
\newblock In {\em The IEEE International Conference on Computer Vision (ICCV)},
  October 2019.

\end{thebibliography}
}

\newpage\phantom{Supplementary}
\multido{\i=1+1}{11}{


\section*{Supplementary material}

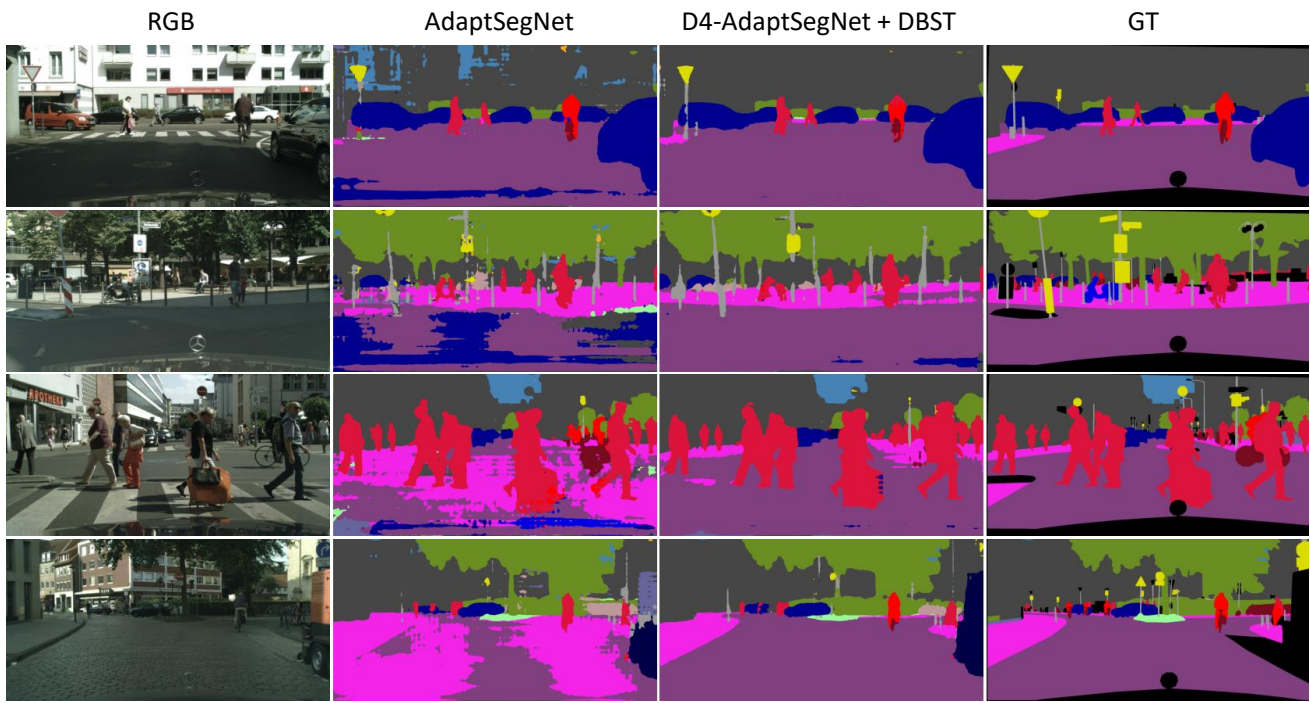

Figure 1. Qualitative results in the GTA5→CS benchmark. From left to right: RGB, prediction from Adaptsegnet [14], prediction from D4-AdaptSegNet + DBST (our proposal), Ground-Truth.

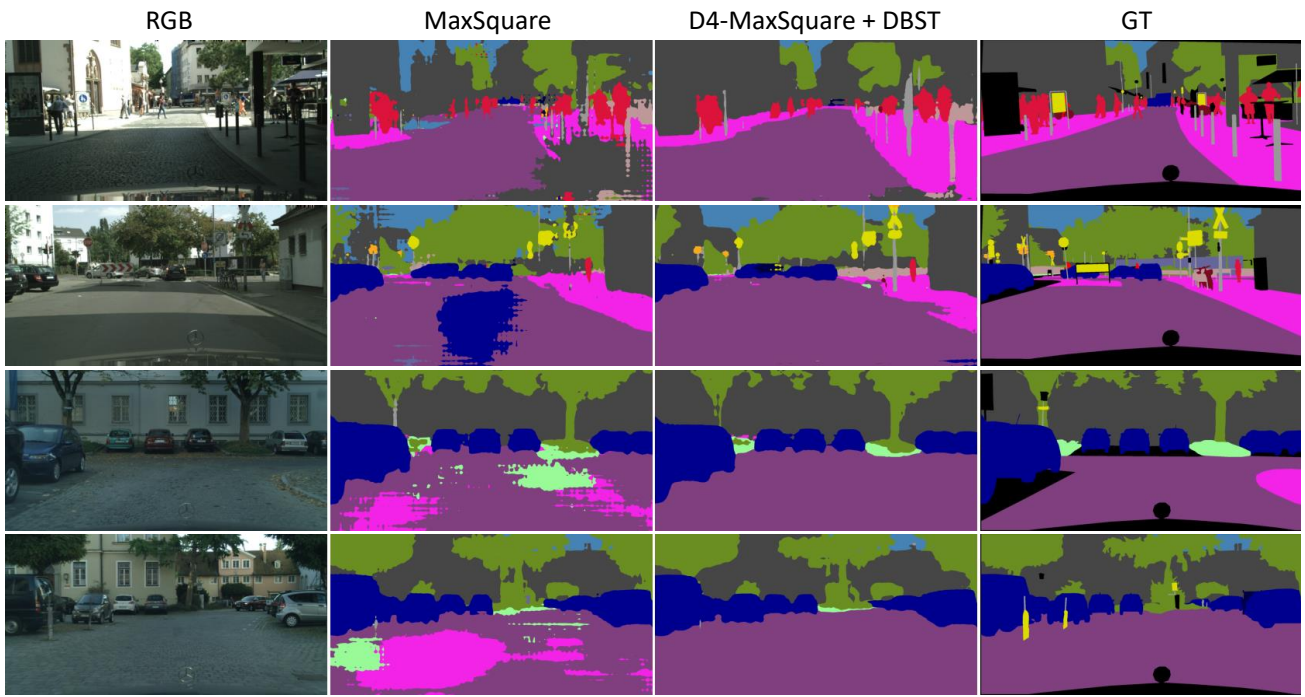

Figure 2. Qualitative results in the GTA5→CS benchmark. From left to right: RGB, prediction from MaxSquare [2], prediction from D4-MaxSquare + DBST (our proposal), Ground-Truth.

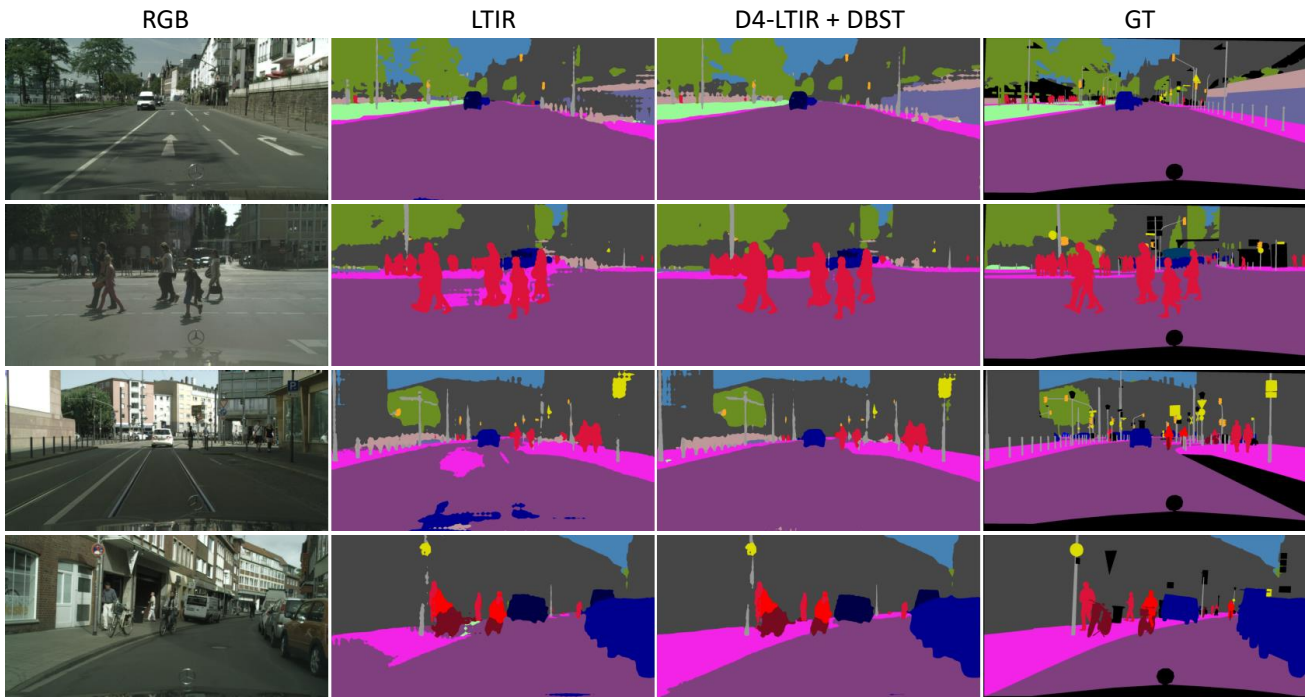

Figure 3. Qualitative results in the GTA5→CS benchmark. From left to right: RGB, prediction from LTIR [7], prediction from D4-LTIR + DBST (our proposal), Ground-Truth.

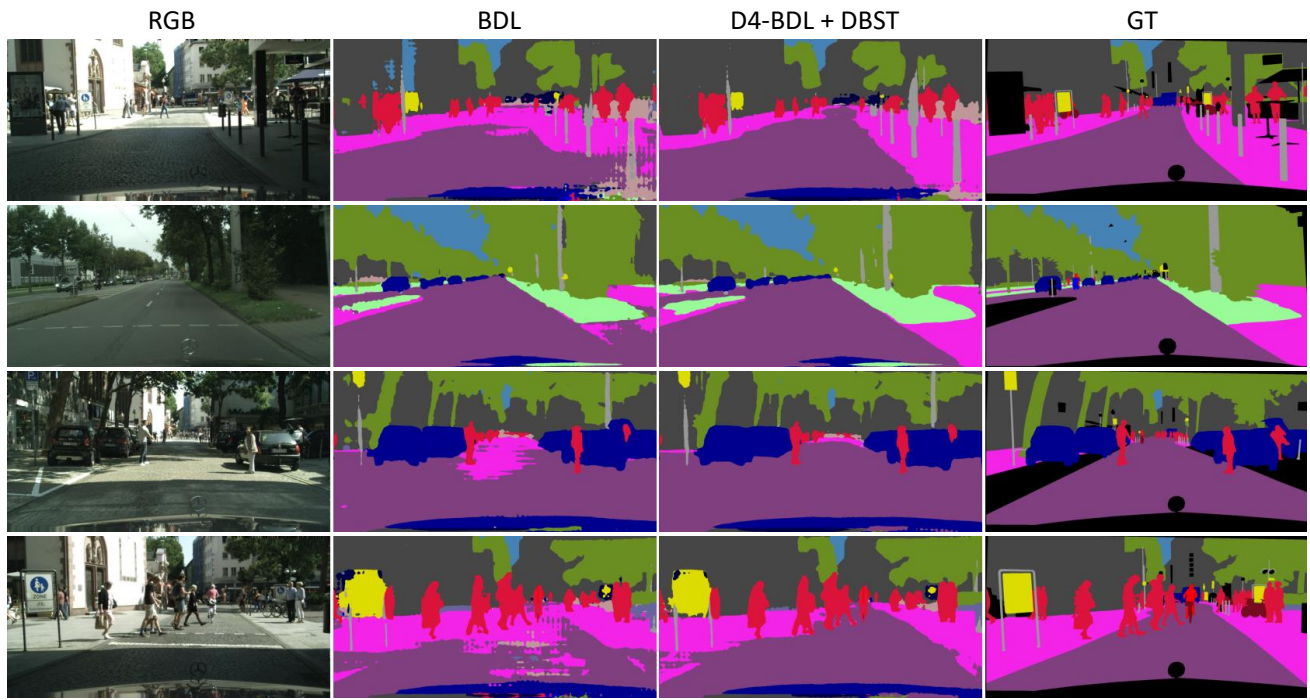

Figure 4. Qualitative results in the GTA5→CS benchmark. From left to right: RGB, prediction from BDL [8], prediction from D4-BDL + DBST (our proposal), Ground-Truth.

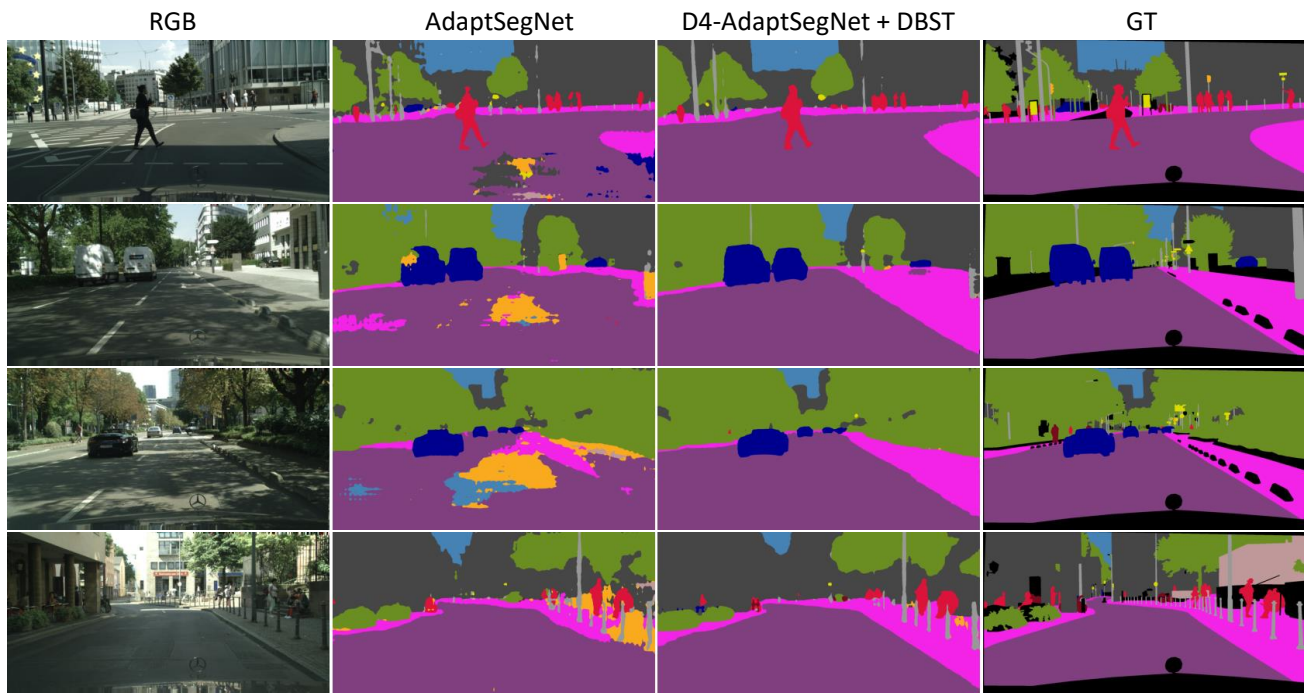

Figure 5. Qualitative results in the SYNSEQ→CS benchmark. From left to right: RGB, prediction from AdaptSegNet [14], prediction from D4-AdaptSegNet + DBST (our proposal), Ground-Truth.

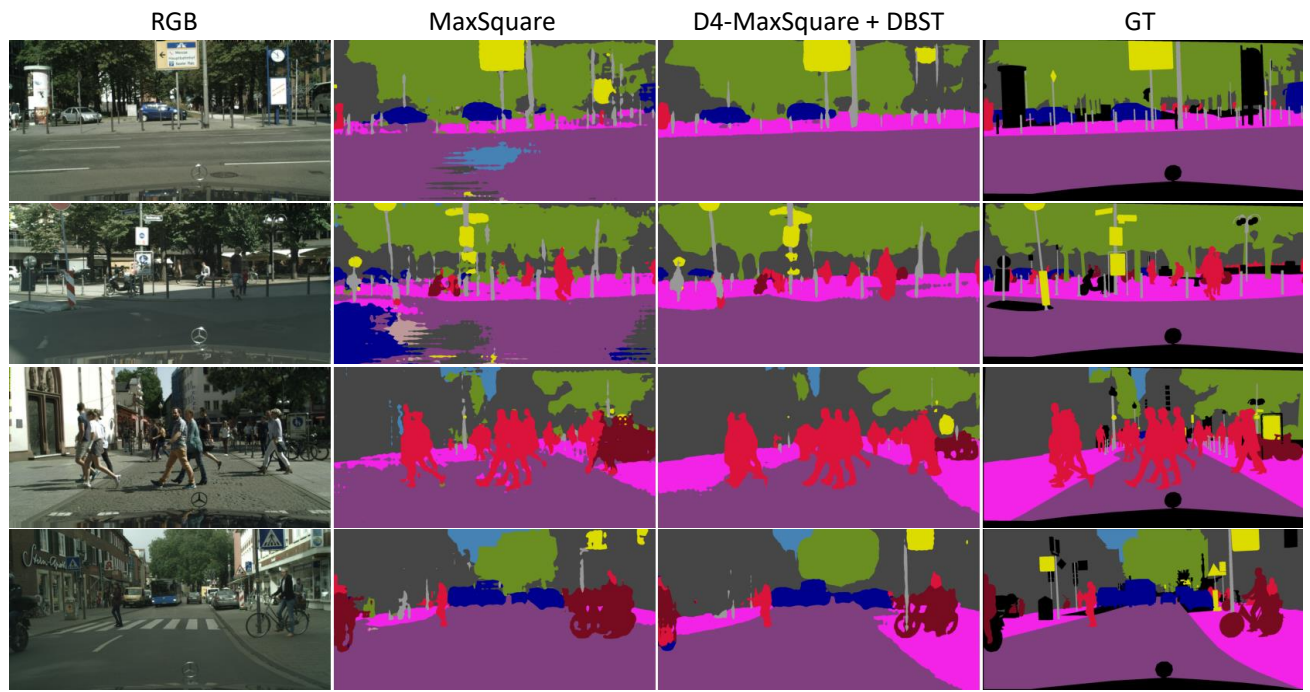

Figure 6. Qualitative results in the SYNSEQ→CS benchmark. From left to right: RGB, prediction from MaxSquare [2], prediction from D4-MaxSquare + DBST (our proposal), Ground-Truth.

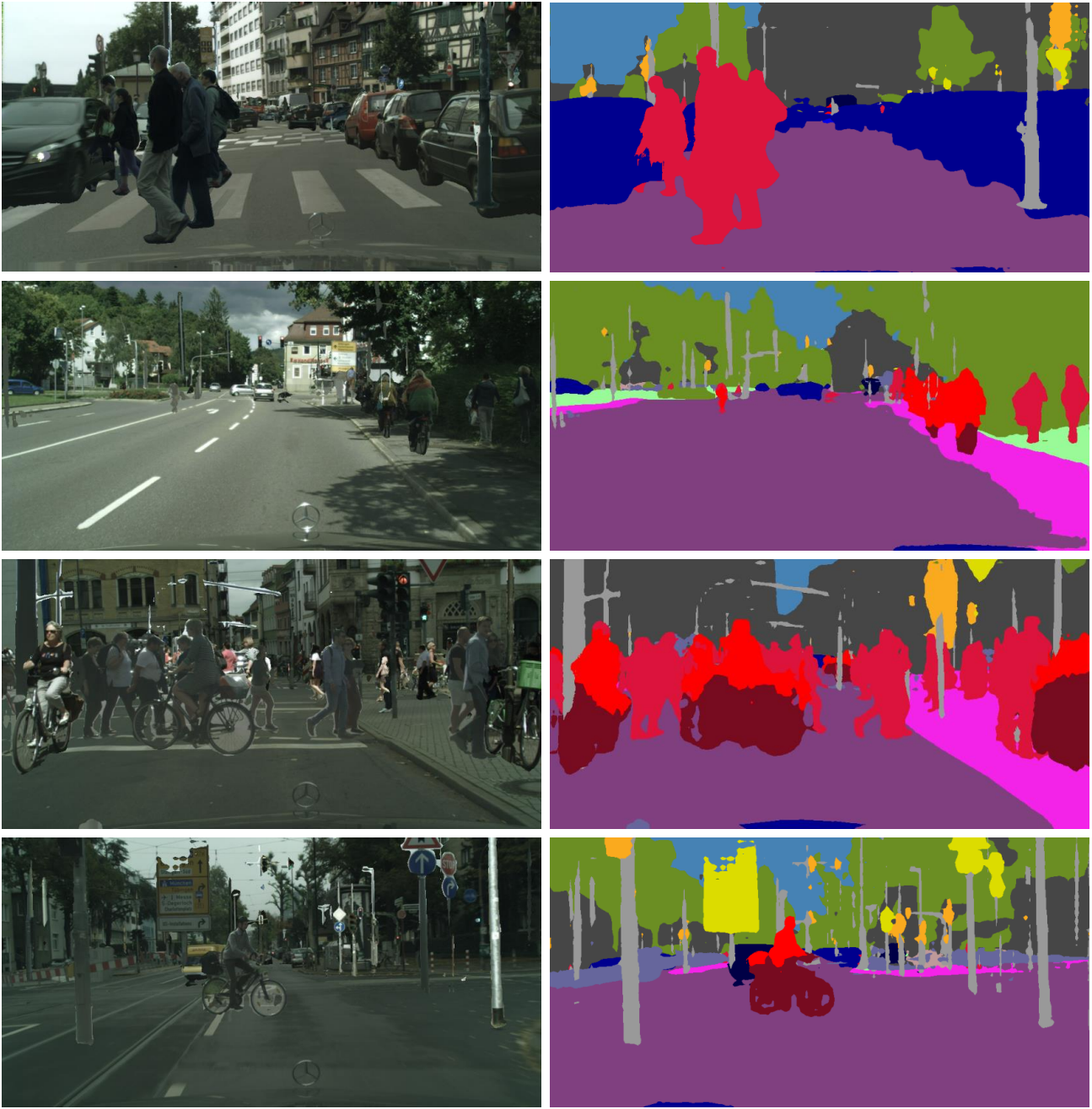

Figure 7. RGB and pseudo-labels generated for our DBST procedure using D4-LTIR in the GTA5→CS benchmark.

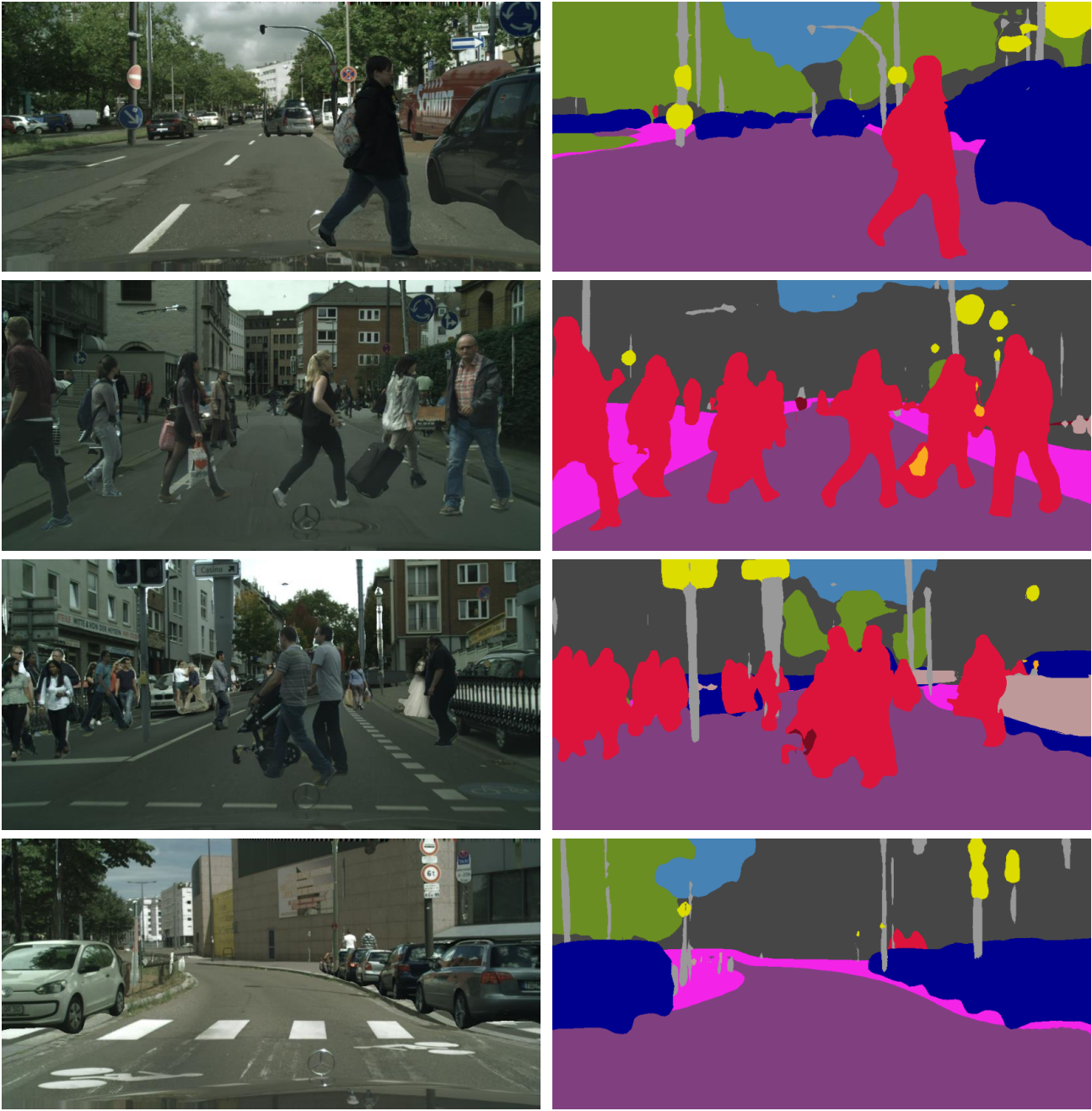

Figure 8. RGB and pseudo-labels generated for our DBST procedure using D4-MRNET in the SYNSEQ→CS benchmark.

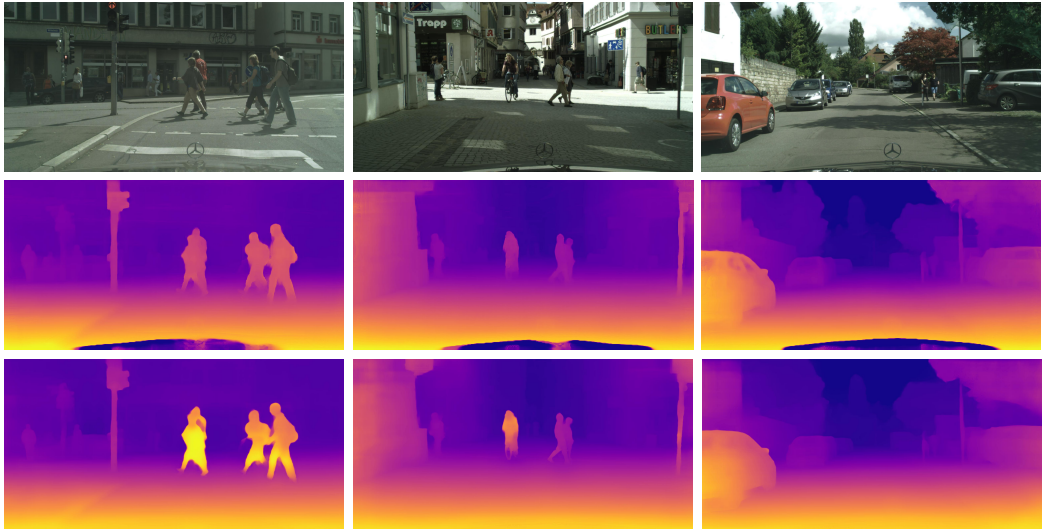

Figure 9. Depth proxy-labels for the Cityscapes dataset obtained with Monodepth2 [6]. From top to bottom: RGB, depth obtained by training Monodepth2 on Cityscapes and GTA5 sequences, depth obtained by training Monodepth2 on Cityscapes and SYNTHIA-SEQ sequences. Depth maps are shown as inverse depth maps for a better visualization.

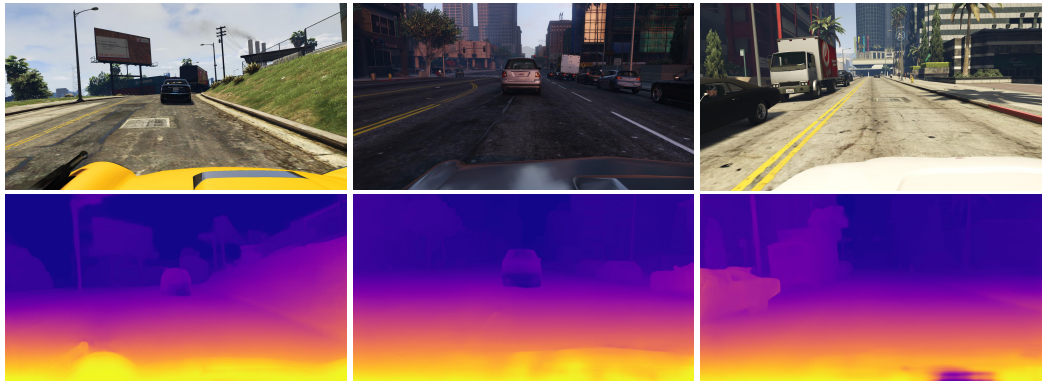

Figure 10. Depth proxy-labels for the GTA5 dataset obtained with Monodepth2 [6]. We show RGB images (first row) and corresponding depth maps (second row), shown as inverse depth maps for a better visualization.

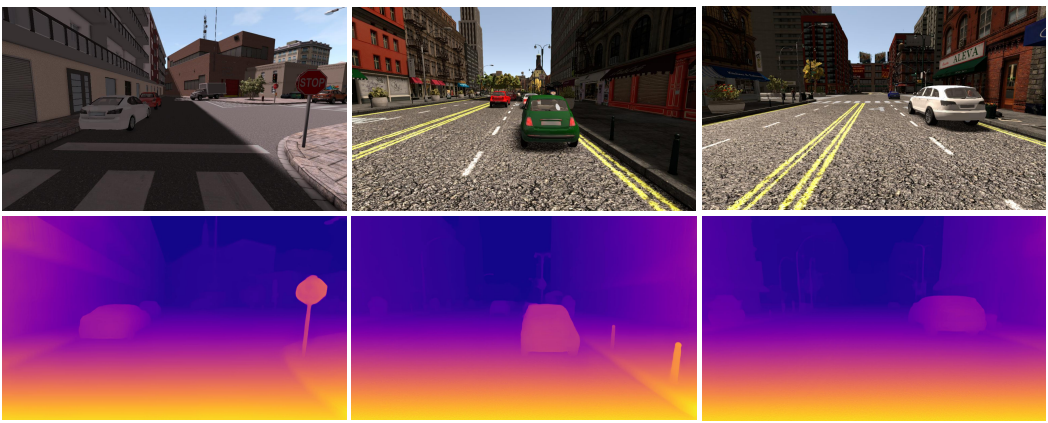

Figure 11. Depth proxy-labels for the SYNTHIA-SEQ dataset obtained with Monodepth2 [6]. We show RGB images (first row) and corresponding depth maps (second row), shown as inverse depth maps for a better visualization.


}

\end{document}